\documentclass[10pt,twocolumn,letterpaper]{article}

\usepackage{iccv}              

\definecolor{iccvblue}{rgb}{0.21,0.49,0.74}

\usepackage[pagebackref,breaklinks,colorlinks,allcolors=iccvblue]{hyperref}
\usepackage[capitalize,noabbrev]{cleveref}
\usepackage{multirow}
\usepackage[noend]{algpseudocode}
\usepackage{amsthm}
\usepackage{microtype}
\usepackage{graphicx}
\usepackage{subcaption}
\usepackage{algorithm}
\usepackage{tikz}
\usepackage{bm}
\PassOptionsToPackage{table}{xcolor}  
\usepackage{xcolor}                   
\usepackage{colortbl} 
\usepackage{booktabs} 
\usepackage{enumitem}

\def\paperID{12043} 
\def\confName{ICCV}
\def\confYear{2025}
\newcommand{\methodname}{SparseVAR} 
\title{Frequency-Aware Autoregressive Modeling for Efficient \\ High-Resolution Image Synthesis}

\author{
    Zhuokun Chen\textsuperscript{\rm 1 \rm 2}\thanks{Email: caesard216@gmail.com} ~~ 
    Jugang Fan\textsuperscript{\rm 1 \rm 2} ~~
    Zhuowei Yu\textsuperscript{\rm 3} ~~
    Bohan Zhuang\textsuperscript{\rm 4}\footnotemark[2] ~~
    Mingkui Tan\textsuperscript{\rm 1 \rm 2}\thanks{Corresponding author. Email: bohan.zhuang@gmail.com , mingkuitan@scut.edu.cn} \\
    \textsuperscript{\rm 1} \small{South China University of Technology} ~~
    \textsuperscript{\rm 2} \small{Pazhou Lab} ~~
    \textsuperscript{\rm 3} \small{University of California, Davis} ~~
    \textsuperscript{\rm 4} \small{Zhejiang University}
}

\begin{document}
\maketitle








\begin{abstract}
Visual autoregressive modeling, based on the next-scale prediction paradigm, exhibits notable advantages in image quality and model scalability over traditional autoregressive and diffusion models. It generates images by progressively refining resolution across multiple stages. However, the computational overhead in high-resolution stages remains a critical challenge due to the substantial number of tokens involved. In this paper, we introduce \methodname, a plug-and-play acceleration framework for next-scale prediction that dynamically excludes low-frequency tokens during inference without requiring additional training. Our approach is motivated by the observation that tokens in low-frequency regions have a negligible impact on image quality in high-resolution stages and exhibit strong similarity with neighboring tokens. Additionally, we observe that different blocks in the next-scale prediction model focus on distinct regions, with some concentrating on high-frequency areas. ~\methodname~leverages these insights by employing lightweight MSE-based metrics to identify low-frequency tokens while preserving the fidelity of excluded regions through a small set of uniformly sampled anchor tokens. By significantly reducing the computational cost while maintaining high image generation quality, ~\methodname~achieves notable acceleration in both HART and Infinity. Specifically, ~\methodname~achieves up to a 2× speedup with minimal quality degradation in Infinity-2B. Code is available at \href{https://github.com/Caesarhhh/SparseVAR.git}{https://github.com/Caesarhhh/SparseVAR}.
\end{abstract}

\section{Introduction}
\label{introduction}

\begin{figure}[t]
  \centering
  \includegraphics[width=0.45\textwidth]{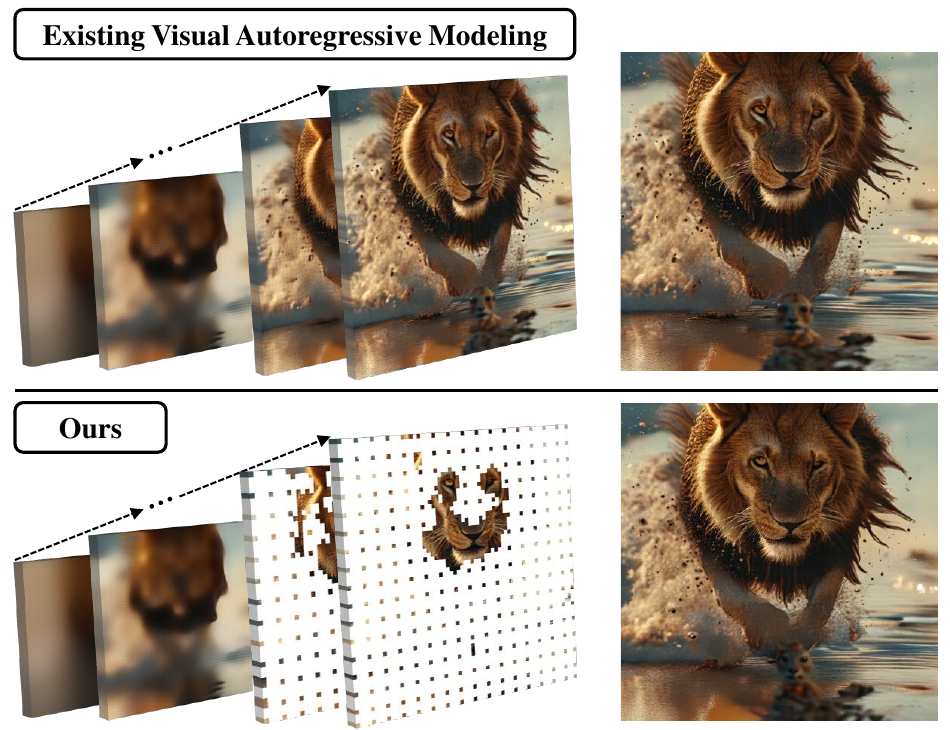}
   \caption{Existing visual autoregressive models allocate uniform computational resources across all regions of a high-resolution image. However, the large number of tokens processed in parallel during high-resolution stages leads to substantial computational overhead. To address this, our method decomposes the target image into high- and low-frequency components, effectively reducing the computational cost in high-resolution stages, thereby lowering inference latency while preserving image generation quality.}
   \label{fig:radar}
\end{figure}

\begin{figure*}[t]
  \centering
    \includegraphics[width=\linewidth]{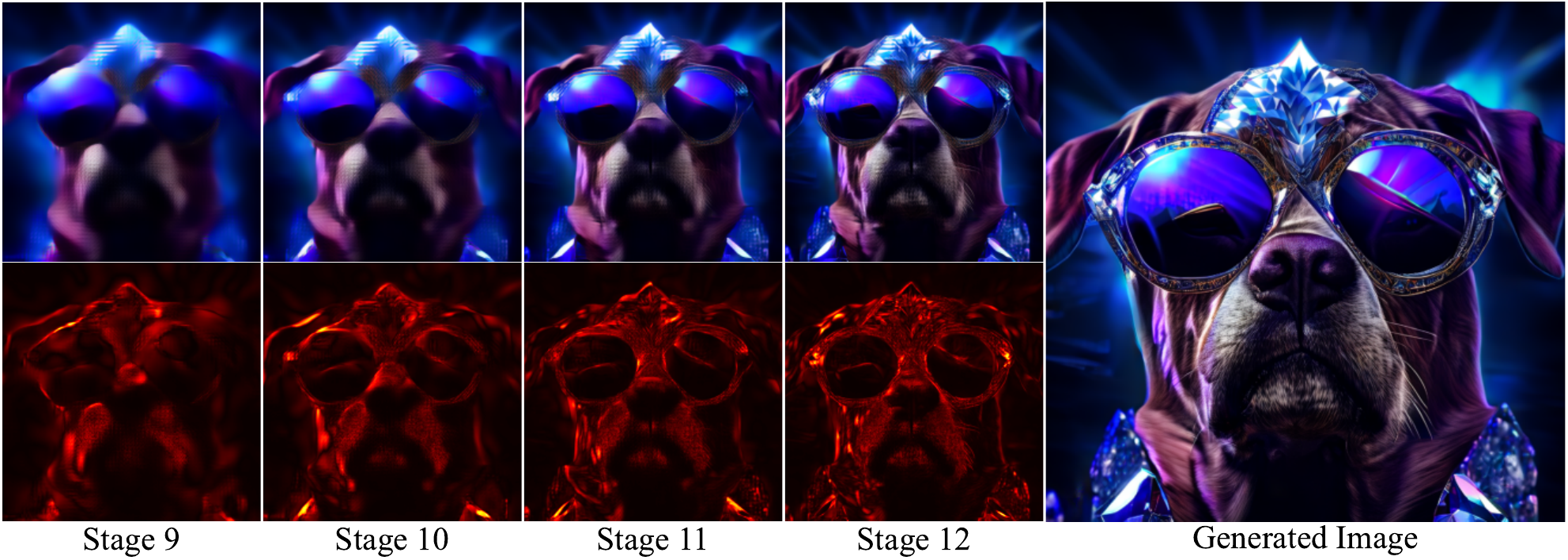}
\vspace{-0.8cm}
\caption{\textbf{High-resolution stages have minimal impact on low-frequency regions.} We visualize the images generated by the last five higher-resolution stages of HART-0.7B (top), along with the $\ell_1$ difference between each stage and its previous stage (bottom). The redder areas indicate where the $\ell_1$ difference is larger, and these areas are mostly concentrated in the high-frequency regions of the foreground. In contrast, the $\ell_1$ change in low-frequency regions, such as the background, is minimal, highlighting that high-resolution stages predominantly focus on high-frequency regions.
}
    \label{fig:motivation}
\end{figure*}

Text-to-image generation has seen widespread application across a range of fields, from creative industries to practical domains like virtual reality and content creation~\citep{reed2016generative,li2019controllable,xu2018attngan,wang2017tag,chen2020optical,rombach2022high,ramesh2021zero}. Among the various approaches, autoregressive models~\citep{wang2024emu3,lee2022autoregressive,mentzerfinite,yulanguage,li2024imagefolder,li2024controlvar,sun2024autoregressive,ding2021cogview} stand out by utilizing a pre-trained tokenizer to quantize continuous image features into a sequence of discrete features by referring to a codebook. This allows the model to generate images by predicting the tokens in a sequence, achieving impressive generalization and scalability. Building on this, next-scale prediction~\citep{tang2024hart,han2024infinity,VAR} further accelerates autoregressive inference by progressively increasing image resolution and predicting the token maps of each resolution stage. This approach generates images in multiple stages, with each stage progressively increasing the resolution. By predicting an entire resolution at each stage, the number of iterations required for high-resolution image generation is significantly reduced. However, during the high-resolution stages, next-scale prediction typically requires the generation of thousands of tokens per resolution, leading to substantial computational overhead, presenting a major challenge for scaling autoregressive models in high-resolution image synthesis.

To reduce the computational burden during high-resolution stages, a natural approach is to decrease the number of tokens involved in the computation. Previous works have extensively explored methods to reduce the number of tokens in vision transformers and multimodal large language models~\citep{bolya2022tome,shang2024llavaprumerge,chen2025fastv,arif2024hired}, which can be broadly categorized into two strategies: merging and selection. Token merging~\citep{bolya2022tome,shang2024llavaprumerge,lee2024vtm} exploit the inherent similarity across visual tokens, using similarity matching or clustering to combine similar tokens. However, 
in high-resolution image generation, the large number of tokens involved makes techniques like clustering and similarity matching computationally prohibitive. Token selection~\citep{chen2025fastv,arif2024hired,he2024zipvl} relies on the redundancy of token attention scores, removing tokens with low attention scores in earlier layers to reduce the token count in subsequent layers, or applying sparse attention operators to accelerate computation. However, as shown in previous work~\citep{he2024zipar,VAR}, autoregressive image generation models exhibit strong local dependencies in token attention scores, where nearly all tokens assign high attention scores to their neighboring tokens and low scores to distant tokens. This consistent pattern across tokens makes it difficult to distinguish redundant tokens based solely on attention scores. Detailed visualizations of the next-scale prediction model's attention map are provided in the appendix.

\begin{figure}[t]
  \centering
    \includegraphics[width=\linewidth]{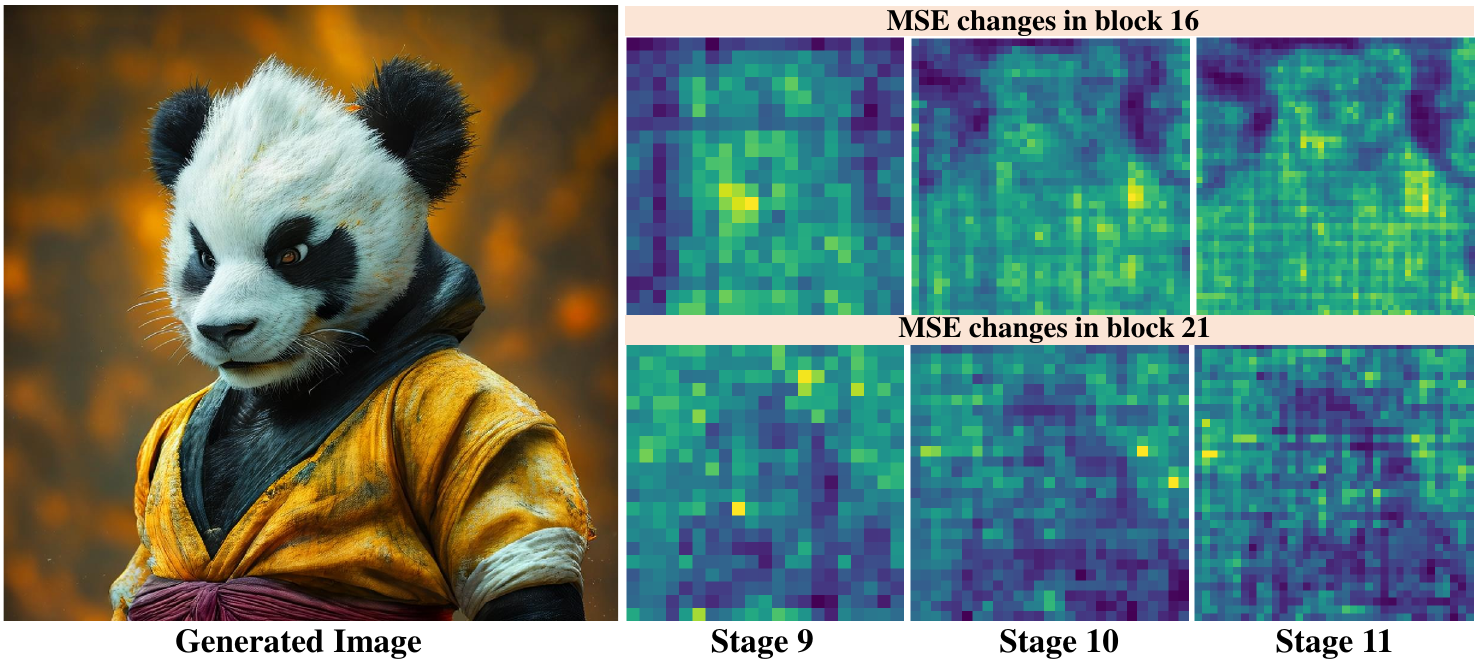}
\vspace{-0.5cm}
\caption{\textbf{Different blocks in next-scale prediction models tend to focus on distinct regions.} We visualize the MSE changes before and after feature inference at the 16th and 21st blocks during stages 10-12 of HART-0.7B. It is clear that different blocks exhibit distinct regional focus tendencies.
}
    \label{fig:motivation2}
\end{figure}

Our analysis begins by exploring the redundancy of token maps across different stages. As shown in Figure~\ref{fig:motivation}, we visualize the $\ell_1$  difference between images generated with and without the residuals from the final several stages of HART-0.7B~\citep{tang2024hart}. The residuals are concentrated in high-frequency regions, while the impact of the final stage’s residuals on low-frequency regions is negligible. This indicates substantial redundancy in token inference during high-resolution stages. Next, as shown in Figure~\ref{fig:motivation2}, we visualize the MSE changes in the features before and after inference across different blocks of the HART-0.7B. Our observations indicate that different blocks attend to distinct regions, with certain blocks exhibiting pronounced changes in high-frequency regions. 

Based on the above observations, we propose \methodname, a plug-and-play method designed to accelerate any next-scale prediction model without the need for additional training. Starting from a relatively high-resolution stage, ~\methodname~dynamically identifies low-frequency tokens using a lightweight metric based on the MSE changes observed across features at specific blocks focusing on high-frequency regions, eliminating the need for computationally expensive similarity matching. Tokens identified as low-frequency are skipped in subsequent inference stages. Moreover, ~\methodname~opts to retain a small number of anchor tokens, which serve as proxies to represent the low-frequency regions, to effectively preserve the generation quality  while incurring minimal additional computation.

We evaluate ~\methodname~on leading high-resolution next-scale prediction methods. The results demonstrate that our method significantly accelerates image generation with virtually no loss in quality. For instance, on the GenEval dataset, ~\methodname~improves the inference speed of Infinity by an average of nearly $2\times$ with a minimal quality degradation in the generated images.

Overall, our contributions are as follows:

\begin{itemize}[leftmargin=*]
\item 
We offer new insights into next-scale prediction models:
(1) A significant amount of redundant tokens exist during inference at high-resolution stages, (2) Different blocks focus on distinct regions.
    
\item We introduce \methodname, a simple yet effective method for accelerating next-scale prediction models. ~\methodname~dynamically identifies low-frequency tokens using a lightweight metric, enabling early exclusion of low-frequency tokens during high-resolution stages, thus significantly reducing computational overhead in low-frequency regions. Moreover, ~\methodname~preserves the generation quality of low-frequency regions by retaining specific anchor tokens. 
 
\end{itemize}

\section{Related Work}

\textbf{Next-scale prediction.}
Next-scale prediction~\cite{VAR,tang2024hart,han2024infinity}, first introduced by VAR~\cite{VAR}, demonstrates the potential of the autoregressive paradigm in image generation, rivaling diffusion transformers~\cite{betker2023dalle3,dustin2024sdxl,chen2024pixart-alpha}. Traditional autoregressive (AR) models~\cite{esser2021vqgan,razavi2019vqvae2,yu2021vitvqgan,lee2022rqtransformer} flatten 2D images into 1D sequences of patch-level tokens. However, the spatial locality inherent in images leads to strong correlations among neighboring patches, which conflicts with the unidirectional dependency assumption in AR modeling and limits both scalability and generalization. VAR~\cite{VAR} addresses this limitation by employing a multi-scale VQ-VAE~\cite{van2017vqvae} to represent images as multi-scale token maps. In this framework, each scale's token map is treated as an autoregressive unit, progressively predicting higher-resolution token maps at each step. While effective, the discrete tokenizer~\cite{van2017vqvae} used in VAR struggles to recover fine-grained image details, imposing an upper bound on generation quality. HART~\cite{tang2024hart} mitigates this issue by introducing a continuous-discrete hybrid tokenizer, significantly improving generation quality at higher resolutions. Inspired by binary vector quantization~\cite{yu2024magvitv2}, Infinity~\cite{han2024infinity} further expands the tokenizer vocabulary and adopts bitwise token prediction, enabling more detailed reconstructions. Despite these advances, these models face challenges related to high computational redundancy, particularly during the last few high-resolution stages of generation.

\noindent\textbf{Token reduction.} Reducing the number of input tokens is a common strategy to enhance computational efficiency. Existing approaches primarily employ token selection~\cite{chen2025fastv, arif2024hired, he2024zipvl} or token merging~\cite{bolya2022tome, lee2024vtm, shang2024llavaprumerge}. FastV~\cite{chen2025fastv} ranks tokens based on their attention scores up to the $K$-th layer and prunes the bottom $R\%$, retaining the remaining tokens for subsequent processing. HiRED~\cite{arif2024hired} addresses high-resolution image inputs by dynamically allocating token budgets per sub-image using shallow-layer attention and selecting the top $N$ patches per sub-image based on deeper-layer [CLS] attention. Similarly, ZipVL~\cite{he2024zipvl} employs an adaptive ratio assignment scheme to discard less critical tokens, thereby compressing the KV cache and accelerating the attention operation. However, token selection methods are unsuitable for generative models due to the high interdependence of tokens. Alternatively, token merging approaches reduce redundancy by combining similar tokens. ToMe~\cite{bolya2022tome} divides tokens into two groups, calculates inter-group similarity, and merges the top $N$ pairs of most similar tokens. VTM~\cite{lee2024vtm} introduces a learnable token merging technique for long-form video inputs, considering both token similarity and saliency. LLava-PruMerge~\cite{shang2024llavaprumerge} integrates selection and merging by initially selecting visual tokens based on [CLS] attention scores, followed by merging using $k$-nearest neighbors. In the context of high-resolution image generation, the sheer volume of tokens significantly amplifies the computational cost of clustering and similarity matching, rendering such techniques infeasible for practical applications.

\begin{figure*}[t]
  \centering
    \includegraphics[width=0.99\linewidth]{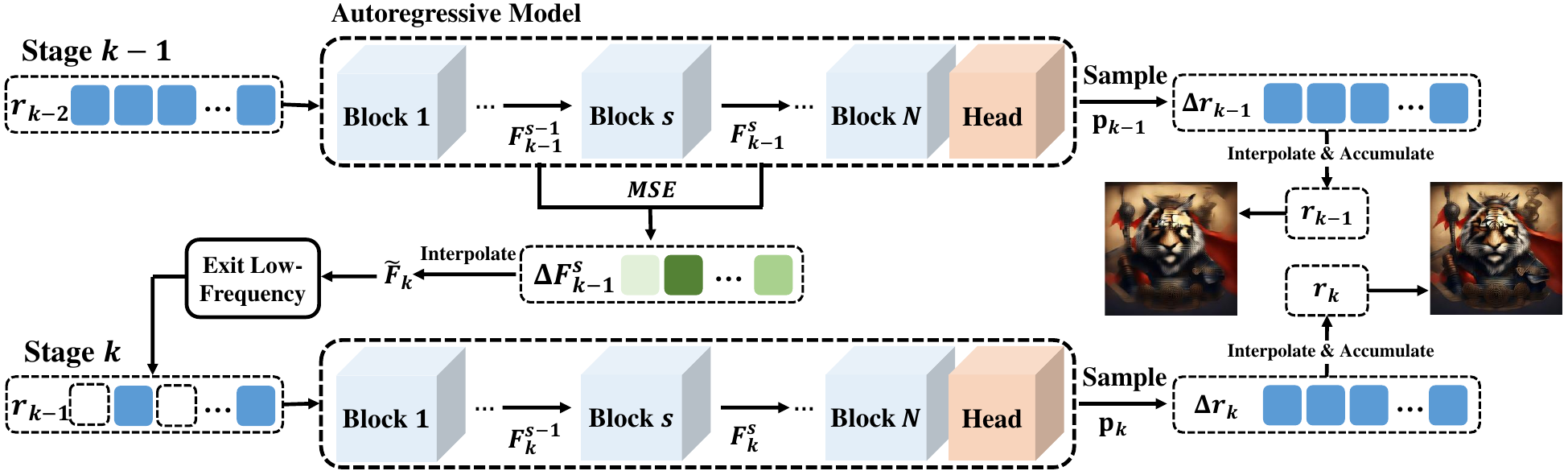}
    \vspace{-0.3cm}
    \caption{Overview of dynamic exclusion in \methodname. ~\methodname~dynamically identifies and excludes low-frequency tokens starting from higher-resolution stages by analyzing MSE changes in features before and after inference in specific blocks, which significantly reduces computational overhead while maintaining generation quality of high-resolution regions.}
    \label{fig:method_exit}
\end{figure*}

\begin{figure}[t]
  \centering
    \includegraphics[width=0.95\linewidth]{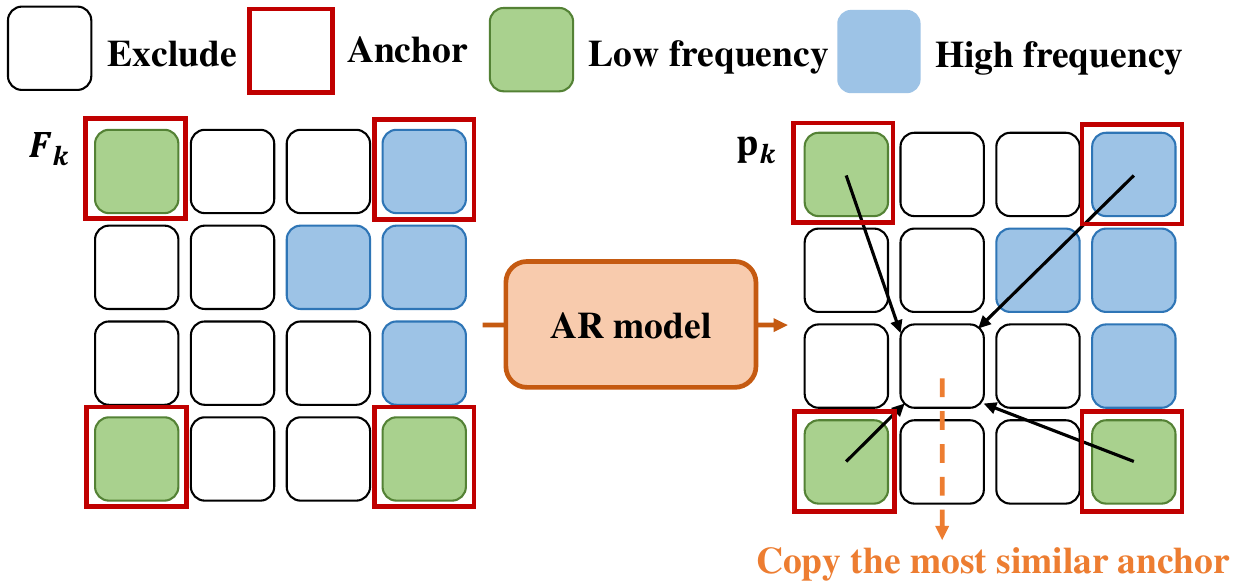}
    \vspace{-0.4cm}
    \caption{Overview of retention of anchor tokens in ~\methodname~($\alpha=3$). ~\methodname~retains a small number of anchor tokens to represent low-frequency regions, enabling early-excluded tokens to copy predictions from the most similar anchors.}
    \label{fig:method_anchor}
\end{figure}

\section{Empircal Insights}
\label{pilot_study}
In this section, we provide visualizations and a in-depth analysis of next-scale prediction models, revealing two key properties that offer critical insights for our method.

\noindent\textbf{Observation 1: The residuals generated at high-resolution stages have minimal impact on low-frequency regions.} Existing VAR models predict logits $\mathbf{\tilde{p}}_k$ at each stage $k$, which are then mapped to residual feature maps $\Delta r_k$ via the pre-trained codebook. To investigate the influence of these predictions on the final image, especially at high-resolution stages, we visualize the $\ell_1$ changes between decoded images from two adjacent stages. As shown in Figure~\ref{fig:motivation}, with increasing stages, the residuals concentrate on high-frequency regions, while their effect on the majority of low-frequency regions is minimal. This indicates significant redundancy in the high-resolution stages. Inspired by this, \textit{we propose early exclusion of low-frequency tokens at high-resolution stages to reduce this redundancy.}

\noindent\textbf{Observation 2: Different blocks in next-scale prediction models tend to focus on distinct regions.} To investigate the regional differences in focus across blocks, we visualize the MSE changes in features before and after inference at various blocks. As shown in Figure~\ref{fig:motivation2}, the regions attended 
differ significantly across blocks. Specifically, the block 16 focuses more on high-frequency regions like contours, while block 21 emphasizes low-frequency regions such as background. Based on this observation, \textit{we propose dynamically distinguishing high- and low-frequency regions of the generated image using the MSE changes in features before and after inference in specific blocks.}



\section{Methodology}

Inspired by the above observations, we propose \methodname, a simple yet effective method for accelerating next-scale prediction models. ~\methodname~comprises two key components: early exclusion of low-frequency tokens and the retention of anchor tokens. As shown in Figure~\ref{fig:method_exit}, Figure~\ref{fig:method_anchor} and Algorithm~\ref{alg:\methodname_stage}, ~\methodname~identifies low-frequency tokens by analyzing the MSE changes at specific blocks, enabling their early exclusion to reduce computational cost. Simultaneously, ~\methodname~retains a set of anchor tokens to ensure the preservation of generation quality in low-frequency regions. The acceleration provided by ~\methodname~is plug-and-play, making it compatible with any next-scale prediction model without the need for additional training.

\subsection{Preliminary}

\textbf{Inference of the next-scale prediction.} Consider a next-scale prediction model comprising $N$ blocks, the VAR framework employs a hierarchical generation process across $K$ progressive resolution scales. At each scale $k \in \{1,...,K\}$, the model parallelly predicts logits $\mathbf{p}_k$ for all $h_k \times w_k$ tokens in the current resolution scale. Subsequently, the residual feature map $\Delta r_k$ is generated by retrieving features for each token from the pretrained codebook based on the predicted logit map $\mathbf{p}_k$. 
The residual feature $\Delta r_i$ from all previous stages ($i \leq k$) are interpolated and accumulated to form $r_{k}$, which serves as the input for stage $k+1$. 
Finally, $r_K$ is used to generate the final image through a VAE decoder.

\subsection{Dynamic Exclusion of Low-Frequency Tokens}

\textbf{Exclusion in high-resolution stages.} As illustrated in \textbf{Observation 1}, we propose to exclude low-frequency tokens during inference to reduce computational overhead. Considering that earlier stages of the next-scale prediction model have lower computational overhead and mainly capture low-frequency information (hence exhibiting limited redundancy), we only start applying early-exit from the \(P\)-th stage onward.

\noindent\textbf{Dynamic high-low frequency identification.} Since the proportion of low-frequency regions varies across images, it is essential to design a lightweight method to dynamically identify regions that are low-frequency and should be excluded from the computation. Inspired by \textbf{Observation 2}, we directly leverage the MSE variations of features within a specific block, to effectively distinguish high- and low-frequency regions. As illustrated in Figure~\ref{fig:method_exit}, let \( \mathbf{F}_{k}^{s} \in \mathbb{R}^{h_k \times w_k \times C} \) represents the output feature map of selected block $s$ at stage \( k(k\geq P)\), where \( C \) is the number of channels. The MSE change map \( \Delta \mathbf{F}_{k-1}^{s} \in \mathbb{R}^{h_{k-1} \times w_{k-1}} \) in previous stage is defined as:

\begin{equation}
\label{eq:mse}
\Delta \mathbf{F}^{s}_{k-1}(i, j) = \frac{1}{C} \sum_{c=1}^{C} \left( \mathbf{F}^{s}_{k-1}(i, j, c) - \mathbf{F}^{s-1}_{k-1}(i, j, c) \right)^2,
\end{equation}
where \( i \) and \( j \) index the spatial dimensions of the feature map. \( \Delta \mathbf{F}_{k-1}^s \) is interpolated to match the resolution of the stage \( k \), resulting in \( \tilde{\mathbf{F}}_{k} \in \mathbb{R}^{h_{k} \times w_{k}} \).

Let \( \mathcal{M}_k^{\text{low}} \subseteq \{1, \dots, h_k\} \times \{1, \dots, w_k\} \) denote the set of low-frequency token indices at stage \( k \). Tokens in stage \( k \) are classified as low-frequency and added to the exclusion set \( \mathcal{M}_k^{\text{low}} \) if their corresponding values in \( \tilde{\mathbf{F}}_k \) are below a threshold \( \tau \cdot \max(\tilde{\mathbf{F}}_k) \), where \( \tau \in [0, 1] \) is a hyperparameter controlling the sparsity level. Formally, we define:
\begin{equation}
\label{eq:exit}
\mathcal{M}_k^{\text{low}} = \{(i, j) \mid \tilde{\mathbf{F}}_{k}(i, j) < \tau \cdot \max(\tilde{\mathbf{F}}_{k})\}.
\end{equation}
Regions identified as low-frequency are excluded from computation in stage \(k\) and all subsequent stages, ensuring that they do not participate in any further computations. 


\subsection{Retention of Anchor Tokens}
To ensure the generation quality of low-frequency regions, we propose retaining a set of anchor tokens that encapsulate the essential information of excluded low-frequency regions by leveraging the high similarity of tokens in neighboring low-frequency areas. Specifically, we uniformly select the top-left corner of every \( \alpha \times \alpha \) grid as the anchor token, ensuring efficient representation while preserving essential structural information.
As illustrated in Figure~\ref{fig:method_anchor}, when \( \alpha = 3 \), anchor tokens are selected from the top-left corner of every \( 3 \times 3 \) grid.

At stage $k$ ($k \geq P$), the output logits map $ \mathbf{p}_{k-1} $ from the previous stage is utilized to assess the similarity of low-frequency regions. Specifically, \( \mathbf{p}_{k-1} \) is interpolated to match the resolution of stage $k$, resulting in \( \mathbf{\tilde{p}}_k \). The logits of anchor tokens, denoted as \( \mathbf{a}_k \), are a subset of \( \mathbf{\tilde{p}}_k \). The cosine similarity between \( \mathbf{\tilde{p}}_k \) and \( \mathbf{a}_k \) is computed as:

\begin{equation}
\label{eq:sim_logits}
\text{Sim}(\mathbf{\tilde{p}}_k, \mathbf{a}_k) = \frac{\mathbf{\tilde{p}}_k^\top \mathbf{a}_k}{\|\mathbf{\tilde{p}}_k\| \|\mathbf{a}_k\|}.
\end{equation}

After inference in stage $k$, an excluded token is assigned the logits of its most similar anchor token if the similarity exceeds a predefined threshold \( \beta \); otherwise, the feature map $\Delta r_k$ in its location is set to zero.



\noindent\textbf{Complexity analysis.}  
\textit{For inference complexity, ~\methodname~reduces the computational cost by excluding low-frequency tokens starting from stage $P$.}  
Since the computational cost of higher-resolution stages dominates the overall inference complexity in next-scale prediction models, we primarily analyze these stages.

For the $k$-th stage, the computational cost of the original model is $O\left(h_k^2 \cdot w_k^2\right)$. ~\methodname~dynamically identifies the low-frequency regions, and suppose a proportion $s_k$ of low-frequency tokens is excluded from computation. Additionally, anchor tokens are uniformly sampled from the feature map at each stage. The number of anchor tokens is proportional to $\frac{1}{\alpha^2}$, where $\alpha$ is the sampling size. The reduced computational cost at stage $k$, considering both the exclusion of low-frequency tokens and the inclusion of anchor tokens, becomes:
\[
O\left( (1 - s_k + \frac{1}{\alpha^2})^2 \cdot h_k^2 \cdot w_k^2 \right).
\]

\begin{algorithm}[ht]
\caption{Inference Procedure for Stage $k$ in \methodname}
\label{alg:\methodname_stage}
\textbf{Input:} Input feature map $r_{k-1}$, selected block index $s$ for MSE computation, threshold $\tau$, anchor grid size $\alpha$, similarity threshold $\beta$, $[\mathbf{p}_{k-1}, \Delta \mathbf{F}^s_{k-1}]$ \text{ (for $k \ge P$)} \\
\textbf{Output:} Logits $\mathbf{p}_k$, MSE change map $\Delta \mathbf{F}_{k}^s$(for $k \ge P-1$)
\begin{algorithmic}[1]
        \If{$k < P$}
            \State Directly inference and obtain logits $\mathbf{p}_k$
            \If{$k = P-1$}
                    \State Compute MSE change map $\Delta \mathbf{F}^s_k$ using Eq.~\eqref{eq:mse}
            \EndIf
        \Else
            \Statex \hspace{1.5em} // Exclude low-frequency tokens dynamically
                \State Interpolate $\Delta \mathbf{F}^s_{k-1}$ to resolution $\mathbf{\tilde{F}}^s_k$
                \State Identify low-frequency tokens $\mathcal{M}_k^{\text{low}}$ using Eq.~\eqref{eq:exit}
                \State Exclude tokens in $\mathcal{M}_k^{\text{low}}$ except anchor tokens
                \State Inference and obtain logits $\mathbf{p}_k$ for remaining tokens

            \Statex \hspace{1.5em} // Copy logits from anchor token
                \State Interpolate logits $\mathbf{p}_{k-1}$ to match resolution the $k$-th stage and obtain $\mathbf{\tilde{p}}_k$
                \For{each token $(i, j) \in \mathcal{M}_k^{\text{low}}$}
                        \State Compute similarity with anchors using Eq.~\eqref{eq:sim_logits}
                        \If{maximum similarity $\geq \beta$}
                                \State Assign logits of the most similar anchor
                        \EndIf
                \EndFor

            \State Compute MSE change map $\Delta \mathbf{F}^s_k$ for the current stage using Eq.~\eqref{eq:mse}
        \EndIf
        \State \Return $\mathbf{p}_k$, $\Delta \mathbf{F}^s_k$(for $k \ge P-1$)
\end{algorithmic}
\end{algorithm}




\section{Experiments}
\label{experiments}

\textbf{Implementation details.} We conduct experiments on the Infinity-2B~\citep{han2024infinity} and HART-0.7B~\citep{tang2024hart}, both are $1024 \times 1024$ high-resolution autoregressive generation models based on next-scale prediction. We compare the performance of accelerated image generation on the GenEval~\citep{ghosh2024geneval}, DPG-Bench~\citep{hu2024ella}, ImageReward~\citep{xu2024imagereward}, and HPSV2.1~\citep{wu2023hps} datasets with $\beta=0.9$ and $P=10$. Since the sparsity varies dynamically across images, we present the average per-image inference performance across these datasets. All inference latency is measured on an NVIDIA 3090 GPU.

\begin{table*}[h]
\centering
\caption{Quantitative evaluation on GenEval. This table presents a comprehensive quantitative analysis of the GenEval benchmark, accounting for varying thresholds $\tau$ and $\alpha=4$. Latency measurements were conducted with a batch size of 1 on a single GPU. The evaluation of Infinity-2B was performed using rewritten prompts in accordance with the methodology outlined in the official repository.}
\vspace{-0.2cm}
\begin{tabular}{cccccccccc}
\toprule[1.5pt]
\multirow{2}{*}{\textbf{Model}} & \multirow{2}{*}{\bm{$\tau$}} & \multicolumn{7}{c}{\textbf{GenEval}$\uparrow$} & \multirow{2}{*}{\textbf{Latency (s)}$\downarrow$} \\ 
\cmidrule(lr){3-9}
                        &      & \textbf{Two Obj.}    & \textbf{Position}   & \textbf{Color} \textbf{Attri.}   & \textbf{Counting}  & \textbf{Colors}   & \textbf{Sin Obj.} & \textbf{Overall}   \\ \midrule
\rowcolor{gray!20}
\textbf{Infinity-2B}                   & - & 0.8586 & 0.4175 & 0.5525 & 0.6844 & 0.8431 & 1.0000 & 0.7260 & 2.78 \\
\midrule
\multirow{4}{*}{\textbf{+ \methodname}} & 0.4 & 0.8485 & 0.4250 & 0.5625 & 0.7000 & 0.8457 & 1.0000 & 0.7303 & 2.64 \\
 & 0.5 & 0.8359 & 0.4250 & 0.5600 & 0.6781 & 0.8351 & 1.0000 & 0.7224 &  1.87 \\
 & 0.6 & 0.8409 & 0.4125 & 0.5475 & 0.6812 & 0.8404 & 1.0000 & 0.7204 & 1.47  \\
 & 0.7 & 0.8460 & 0.4225 & 0.5475 & 0.6719 & 0.8378 & 1.0000 & 0.7209  & 1.36  \\
 \midrule
 \rowcolor{gray!20}
 \textbf{HART-0.7B}                   & - & 0.6919 & 0.1625 & 0.2825 & 0.3688 & 0.8617 & 0.9938 & 0.5602 & 1.32 \\\midrule
\multirow{4}{*}{\textbf{+ \methodname}} & 0.4 & 0.7071 & 0.1450 & 0.2650 & 0.3938 & 0.8777 & 0.9906 & 0.5632 & 1.25 \\
 & 0.5  & 0.7045 & 0.1600 & 0.2575 & 0.3969 & 0.8644 & 0.9906 & 0.5623 & 1.18 \\
 & 0.6 & 0.7071 & 0.1600 & 0.2825 & 0.3562 & 0.8670 & 0.9906 & 0.5606 & 0.99 \\
 & 0.7 & 0.6035 & 0.1200 & 0.2125 & 0.3344 & 0.8351 & 0.9656 & 0.5119 & 0.81 \\
\bottomrule[1.5pt]
\end{tabular}
\label{tab:results_infinity_geneval}
\end{table*}

\noindent\textbf{Main results.} To evaluate the acceleration performance of \methodname, we conduct experiments on the GenEval and DPG-Bench. As shown in Table~\ref{tab:results_infinity_geneval} and Table~\ref{tab:results_dpgbench}, the results demonstrate that ~\methodname~significantly improves inference speed with minimal impact on image generation quality. For instance, when $\tau = 0.7$, ~\methodname~reduces the inference latency of Infinity by 51\%, with the overall score decreasing by only 0.0051 in GenEval and 0.0033 in DPG-Bench. Similarly, when $\tau = 0.6$, ~\methodname~achieves a 25\% reduction in the inference latency of HART, while exhibiting only a marginal decrease in GenEval(0.0030) and in DPG-Bench(0.0023). These results indicate that ~\methodname~effectively preserves the image generation quality of high-frequency regions during high-resolution stages.

\begin{table}[h]
\centering
\renewcommand{\tabcolsep}{1.8pt}
\caption{Quantitative evaluation on DPG-Bench. Latency measurements are conducted on a single GPU using a batch size of 1.}
\vspace{-0.1cm}
\begin{tabular}{cccccc}
\toprule[1.5pt]
\multirow{2}{*}{\textbf{Model}} & \multirow{2}{*}{\bm{$\tau$}} & \multicolumn{3}{c}{\textbf{DPG-Bench}$\uparrow$} & \multirow{2}{*}{\textbf{Latency (s)}$\downarrow$} \\ 
\cmidrule(lr){3-5}
                        &      & \textbf{Global.}    & \textbf{Relation} & \textbf{Overall}   \\ \midrule
\rowcolor{gray!20}
\textbf{Infinity-2B}                   & - & 0.8419 & 0.9283 & 0.8289 & 2.55 \\
\midrule
\multirow{4}{*}{\textbf{+ \methodname}} & 0.4 & 0.8541 & 0.9246 & 0.8282 & 2.34  \\
 & 0.5 & 0.8480 & 0.9242 & 0.8254 & 1.69 \\
 & 0.6 & 0.8632 & 0.9237 & 0.8260 & 1.35  \\
 & 0.7 & 0.8511 & 0.9270 & 0.8256 & 1.20  \\
 \midrule
 \rowcolor{gray!20}
 \textbf{HART-0.7B}                   & - & 0.8710 & 0.9295 & 0.8099 & 1.31 \\\midrule
\multirow{4}{*}{\textbf{+ \methodname}} & 0.4 & 0.8571 & 0.9233 & 0.8092 & 1.24 \\
 & 0.5  & 0.8602 & 0.9233 & 0.8082 & 1.19 \\
 & 0.6 & 0.8602 & 0.9246 & 0.8069 & 1.00 \\
 & 0.7 & 0.8663 & 0.9254 & 0.8072 & 0.83 \\
\bottomrule[1.5pt]
\end{tabular}
\label{tab:results_dpgbench}
\end{table}

\noindent\textbf{Human preference evaluation.} To further investigate the impact of \methodname's acceleration on human evaluation preferences, we conducted experiments on two datasets focused on human preference assessment. As shown in Table~\ref{tab:human_preference}, on the HPSv2.1 dataset, when $\tau = 0.7$, ~\methodname~reduces the average inference latency of Infinity by 49.43\% while the overall score decreases by only 0.47. On the ImageReward dataset, ~\methodname~reduces the inference latency by 49.62\% with a score reduction of only 0.0266. These results indicate that ~\methodname~effectively preserves the quality of high-frequency regions, and the generated images remain aligned with human preferences.

\begin{table*}[h]
\centering
\caption{Quantitative evaluation on Human Preference Metrics. This table provides a detailed quantitative analysis on two human preference benchmarks, considering varying thresholds $\tau$ and a fixed local window size of 4. Latency measurements were performed with a batch size of 1 on a single GPU to ensure consistency and accuracy.}
\vspace{-0.2cm}
\begin{tabular}{cccccccccc}
\toprule[1.5pt]
\multirow{2}{*}{\textbf{Model}} & \multirow{2}{*}{\bm{$\tau$}} & \multicolumn{2}{c}{\textbf{ImageReward}} & \multicolumn{6}{c}{\textbf{HPSv2.1}} \\ 
\cmidrule(lr){3-4} \cmidrule(lr){5-10}
                        &      & \textbf{Score}$\uparrow$  & \textbf{Latency(s)}$\downarrow$  & \textbf{Anime}   & \textbf{Concept-Art}   & \textbf{Paintings}  & \textbf{Photo} & \textbf{Overall}$\uparrow$  & \textbf{Latency(s)}$\downarrow$  \\ \midrule
\rowcolor{gray!20}
\textbf{Infinity-2B}                   & - & 0.9212 & 2.64 & 31.63 & 30.26 & 30.28 & 29.27 & 30.36 & 2.61  \\
\midrule
\multirow{4}{*}{\textbf{+ \methodname}} & 0.4 & 0.9147 & 2.37 & 31.58 & 30.13 & 30.16 & 29.22 & 30.27 & 2.35 \\
 & 0.5 & 0.8969 & 1.77 & 31.40 & 29.95 & 29.96 & 29.05 & 30.09 & 1.79 \\
 & 0.6 & 0.8943 & 1.42 & 31.29 & 29.82 & 29.77 & 28.94 & 29.95 & 1.40 \\
 & 0.7 & 0.8946 & 1.33 & 31.21 & 29.75 & 29.71 & 28.88 & 29.89 & 1.32 \\
 \midrule
 \rowcolor{gray!20}
\textbf{HART-0.7B}                   & - & 0.8656 & 1.32 & 31.22 & 29.61 & 29.10 & 28.21 & 29.53 & 1.30  \\
\midrule
\multirow{4}{*}{\textbf{+ \methodname}} & 0.4 & 0.8818 & 1.25 & 31.19 & 29.58 & 29.08 & 28.19 & 29.51 & 1.26 \\
 & 0.5 & 0.8818 & 1.20 & 31.06 & 29.47 & 28.96 & 28.09 & 29.40 & 1.19 \\
 & 0.6 & 0.8121 & 1.01 & 30.25 & 28.68 & 28.13 & 27.51 & 28.64 & 1.02 \\
 & 0.7 & 0.4333 & 0.82 & 27.18 & 25.60 & 25.13 & 24.93 & 25.71 & 0.81 \\
\bottomrule[1.5pt]
\end{tabular}
\label{tab:human_preference}
\end{table*}

\begin{figure*}[h]
    \centering
    \begin{minipage}{0.33\textwidth}
        \centering
        \captionof{table}{Evaluation of different local window sizes $\alpha$ on GenEval. $\tau$ is set as 0.6. Inference latency is measured with batch size of 1 and in seconds. - means that we do not keep anchor tokens.}
        \vspace{-0.3cm}
        \label{tab:window_size_geneval}
        \resizebox{\textwidth}{!}{
        \begin{tabular}{ccccc}
            \toprule[1.5pt]
            \multirow{2}{*}{\bm{$\alpha$}} & \multicolumn{2}{c}{\textbf{Infinity}} & \multicolumn{2}{c}{\textbf{HART}} \\ 
            \cmidrule(lr){2-3} \cmidrule(lr){4-5}
            & \textbf{Score} & \textbf{Latency(s)} & \textbf{Score} & \textbf{Latency(s)} \\ \hline
            2  & 0.7235 & 1.76 & 0.5615 & 1.11 \\
            3  & 0.7210 & 1.54 & 0.5578 & 1.06 \\
            4  & 0.7204 & 1.47 & 0.5606 & 0.99 \\
            5  & 0.7200 & 1.46 & 0.5560 & 0.97 \\
            -  & 0.7190 & 1.38 & 0.5502 & 0.93 \\ 
            \bottomrule[1.5pt]
        \end{tabular}}
    \end{minipage}
    \hfill
    \begin{minipage}{0.33\textwidth}
        \centering
        \captionof{table}{Impact of different $P$ on image generation quality and inference latency on GenEval. We report the generation quality score (↑) and inference latency in seconds (↓) for each $P$.}
        \vspace{-0.2cm}
        \label{tab:influence_of_k}
        \resizebox{\textwidth}{!}{
        \begin{tabular}{ccccc}
            \toprule[1.5pt]
            \multirow{2}{*}{$P$} & \multicolumn{2}{c}{\textbf{Infinity}} & \multicolumn{2}{c}{\textbf{HART}} \\ 
            \cmidrule(lr){2-3} \cmidrule(lr){4-5}
            & \textbf{Score} & \textbf{Latency(s)} & \textbf{Score} & \textbf{Latency(s)} \\ \hline
            6   &  0.6805  & 1.29  & 0.5529 & 0.98 \\
            8   &  0.7085  & 1.35  & 0.5577 & 0.98 \\
            9   &  0.7126  & 1.39  & 0.5565 & 0.99 \\
            10  &  0.7204  & 1.47  & 0.5602 & 0.99 \\
            11  &  0.7261  & 1.76  & 0.5625 & 1.03 \\
            12  &  0.7274  & 2.21  & 0.5607 & 1.05 \\
            \bottomrule[1.5pt]
        \end{tabular}}
    \end{minipage}
    \hfill
    \begin{minipage}{0.3\textwidth}
        \centering
        \vspace{0.2cm}
        \includegraphics[width=\textwidth]{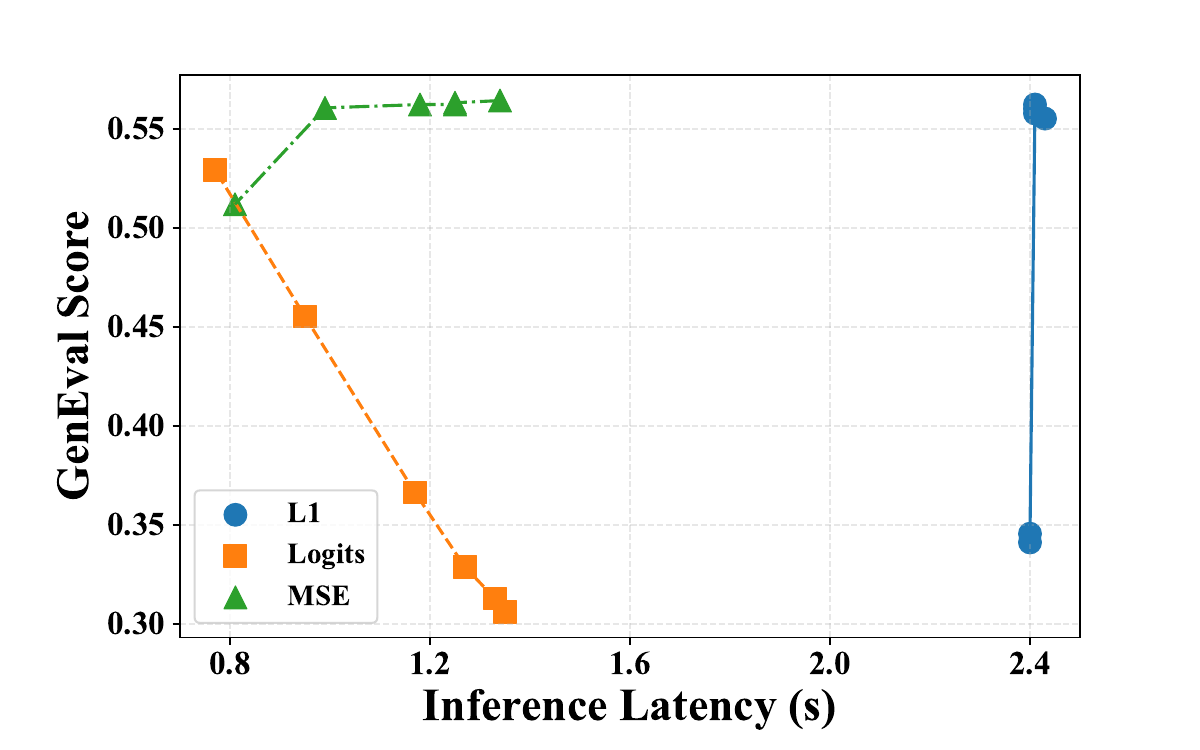}
        \vspace{-0.7cm}
        \caption{Comparison of different metrics for distinguishing high- and low-frequency regions.}
    \label{fig:metrics_comparison}
    \end{minipage}
\end{figure*}

\noindent\textbf{Influence of different $\alpha$.} To investigate the effectiveness of anchor tokens, we conducted comparative experiments on HART and Infinity by varying the grid size $\alpha$. As shown in Table~\ref{tab:window_size_geneval}, retaining anchor tokens improves the image generation quality of both HART and Infinity while introducing minimal additional inference overhead. Notably, HART is more significantly affected due to its residual diffusion mechanism, which takes the final stage's output as input. Without logits assigned to the early exiting low-frequency tokens via anchor tokens, residual diffusion lacks these low-frequency inputs, leading to incomplete detail refinement in these regions. Consequently, the strategy of retaining anchor tokens effectively preserves the optimization quality of this diffusion process. When $\alpha$ is small, the image generation quality improves further, but the number of tokens required for inference increases significantly, resulting in slower inference. Considering both quality and efficiency, we set $\alpha = 4$.

\noindent\textbf{Influence of different $P$.} To investigate the impact of early-exited stages on image generation quality, we conducted experiments on GenEval to compare the effects of different values of $P$ on generation quality and average inference latency. As shown in Table~\ref{tab:influence_of_k}, when $P$ is set to a small value, meaning the model exits at a very early stage, the image generation quality significantly deteriorates while inference latency is not greatly reduced. This is because next-scale prediction progressively increases the resolution, making later stages much more computationally intensive than earlier ones. Furthermore, earlier stages primarily generate low-frequency content, making early exits more sensitive to the final image quality. Therefore, considering both quality and efficiency, we set $P=10$.


\noindent\textbf{Comparisons with different early exiting metrics.} To compare the effectiveness of different metrics for distinguishing high- and low-frequency regions for early exiting, we evaluated three approaches: using the MSE changes in specific block, the logits similarity generated at each stage, and the $\ell_1$ differences between images generated by the cumulative residuals decoded at each stage and the previous stage. We have detailed the calculation specifics of the three metrics in the appendix. As shown in Figure~\ref{fig:metrics_comparison}, using MSE as the metric for early exiting provides the most lightweight and accurate measurement of low-frequency regions, enabling precise early exits. Although using the $\ell_1$ difference of generated images can also effectively distinguish high- and low-frequency regions, it requires an additional decoder operation at each stage, resulting in extra computational overhead.

\begin{figure*}[h]
\centering
\begin{subfigure}[b]{0.99\textwidth}
    \centering
    \includegraphics[width=0.99\linewidth]{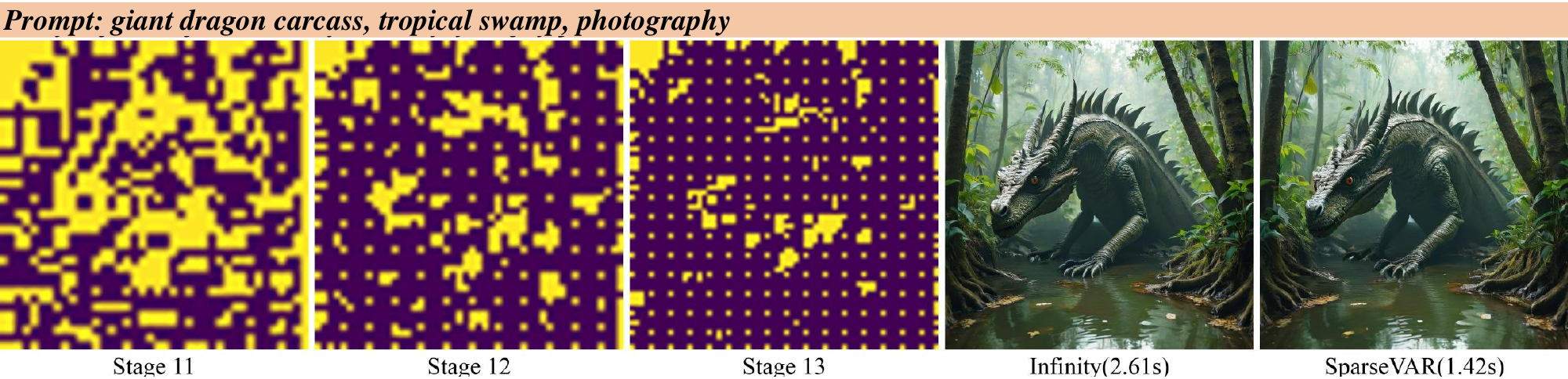}
    \label{fig:vis_sparsity_infinity_1}
\end{subfigure}
\begin{subfigure}[b]{0.99\textwidth}
    \centering
    \includegraphics[width=0.99\linewidth]{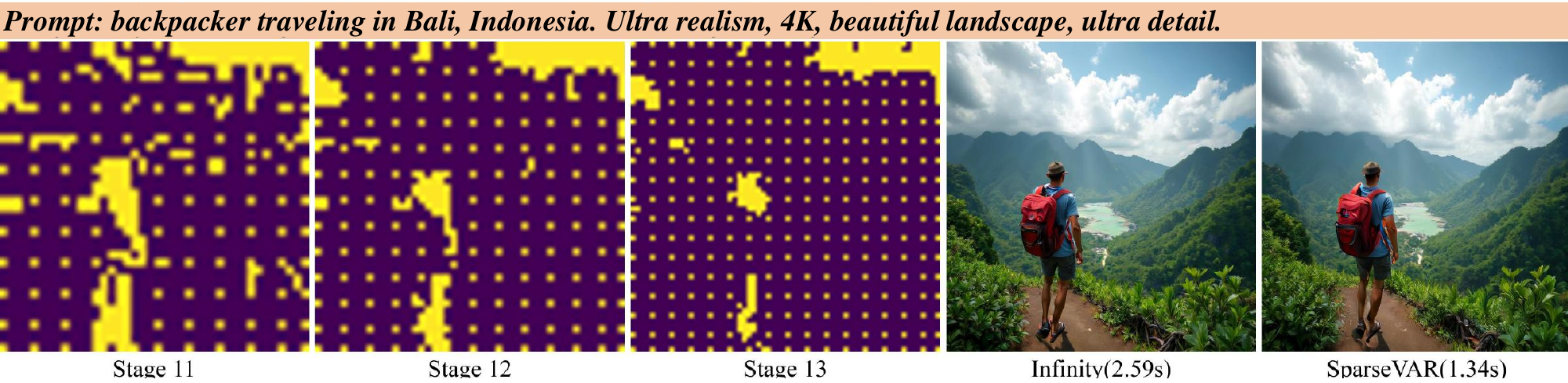}
    \label{fig:vis_sparsity_infinity_3}
\end{subfigure}
\begin{subfigure}[b]{0.99\textwidth}
    \centering
    \includegraphics[width=0.99\linewidth]{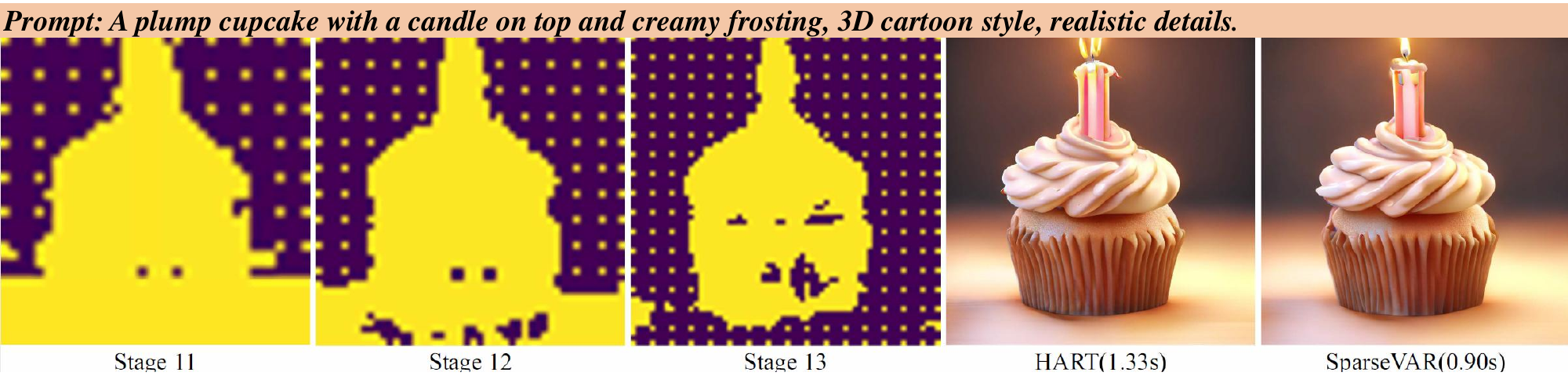}
    \label{fig:vis_sparsity_hart_2}
\end{subfigure}
\vspace{-0.3cm}
\caption{Qualitative visualizations of \methodname. The yellow and purple colors represent the tokens identified as retained and early-exited at each stage, respectively.}
\label{fig:visualizations}
\end{figure*}

\begin{figure}[h]
\centering
\includegraphics[width=0.45\textwidth]{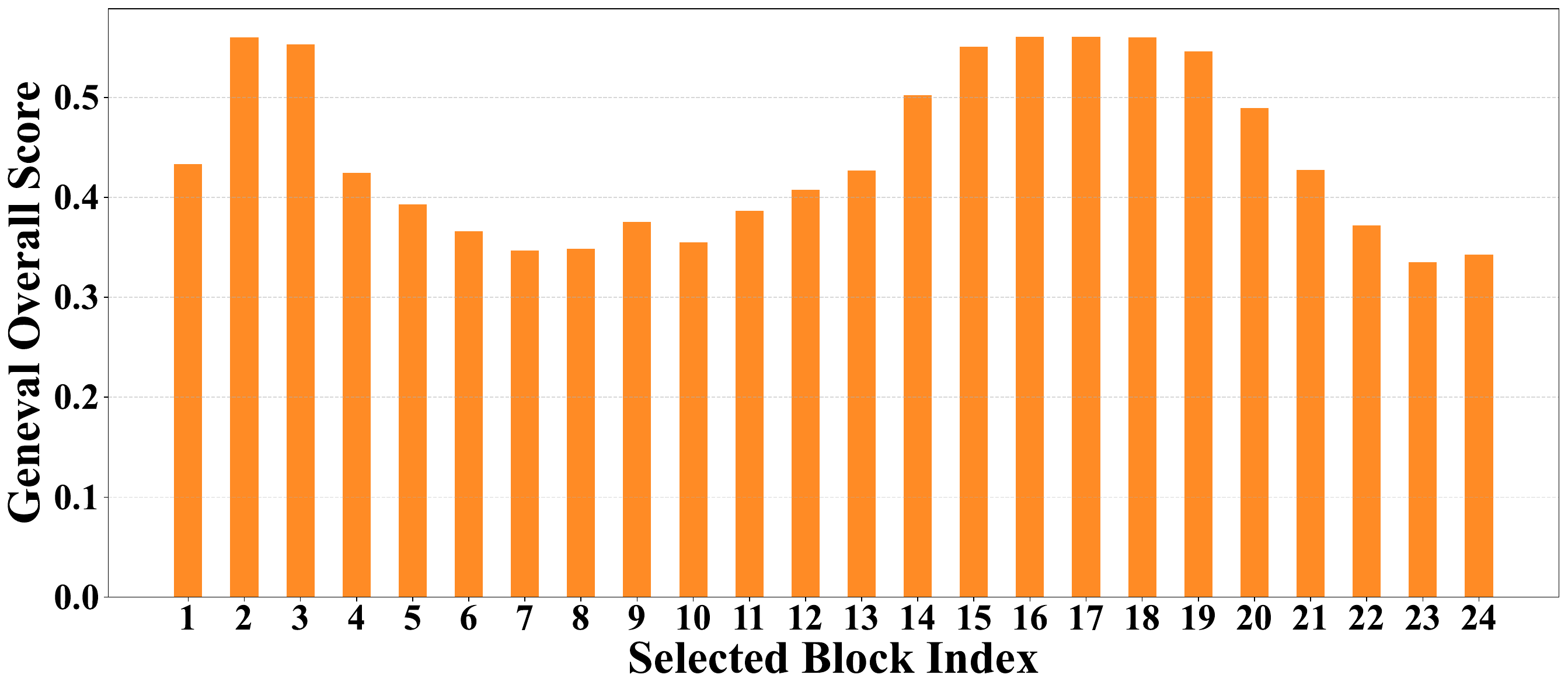}
\vspace{-0.3cm}
\caption{Impact of block selection on MSE-based frequency estimation. Image generation quality is evaluated using the HART-0.7B across different blocks.}
\label{fig:ablation_s}
\end{figure}

\noindent\textbf{Impact of block selection on MSE-based frequency estimation.} To investigate the impact of using MSE changes from different blocks for high- and low-frequency selection on image generation quality, we conducted comparative experiments across various blocks on Infinity with $\tau=0.6$.  As shown in Figure~\ref{fig:ablation_s}, while some blocks exhibit clear distinctions in MSE variations between high and low frequencies, others do not. Among all blocks, the 16$^{th}$ block achieved the best results. Therefore, we utilize the MSE changes from the 16$^{th}$ block for high- and low-frequency estimation. Additional visualizations of MSE changes across different blocks are provided in the appendix.




\noindent\textbf{Qualitative visualizations.} To provide a more intuitive demonstration of \methodname's ability to accurately reduce computations in low-frequency regions, we visualize the token distribution for early exiting across different stages with $\tau = 0.6$ and $\alpha = 4$. As shown in Figure~\ref{fig:visualizations}, the visualizations illustrate the sparsity pattern of \methodname's early exit process in 11$^{th}$ to 13$^{th}$ stage. As clearly illustrated in the figure, ~\methodname~effectively excludes the majority of low-frequency regions, retaining high-frequency regions for inference. By the final stage, only a small number of tokens remain, yet the majority of the image generation quality is preserved. Compared to the baseline, ~\methodname~achieves a significant acceleration in inference speed with minimal degradation in image generation quality. 

\section{Conclusion}
This paper explores the redundancy of token computations in high-resolution stages of next-scale prediction models and proposes \methodname, a novel method for effectively accelerating image generation. The approach offers a simple yet effective strategy that dynamically identifies and excludes low-frequency tokens, requiring no additional training. By leveraging the strong local dependencies between neighboring tokens, ~\methodname~significantly reduces computational overhead while preserving image quality. Overall, ~\methodname~enhances the efficiency of next-scale prediction models with minimal loss in performance, providing a practical solution for high-resolution image synthesis.

\textbf{Limitations and future work.} Our research primarily focuses on next-scale prediction models, leaving the broader applicability of early exclusion of low-frequency tokens in other autoregressive generation models unexplored. Extending this concept to a wider range of autoregressive frameworks remains an important direction for future work. Additionally, the current approach employs a uniform sampling strategy for anchor selection. However, fixed window frequencies and anchor positions may not be optimal for all images, as the characteristics of low-frequency regions vary significantly across different inputs. Developing a dynamic method to adaptively adjust the sampling frequency and anchor placement based on image content could further enhance the generation quality of low-frequency regions.

\section{Acknowledgement}
This work was partially supported by the Joint Funds of
 the National Natural Science Foundation of China (Grant
 No.U24A20327).

\bibliography{example_paper}
\bibliographystyle{icml2025}






%
\definecolor{iccvblue}{rgb}{0.21,0.49,0.74}



\def\paperID{12043} 
\def\confName{ICCV}
\def\confYear{2025}


\newpage
\appendix
\onecolumn
\begin{center}
{
    \Large{\textbf{Appendix}}
}
\end{center}

\section{Performance on Non-Residual VAE Models}

\begin{figure}[h]
  \centering
  \includegraphics[width=0.95\linewidth]{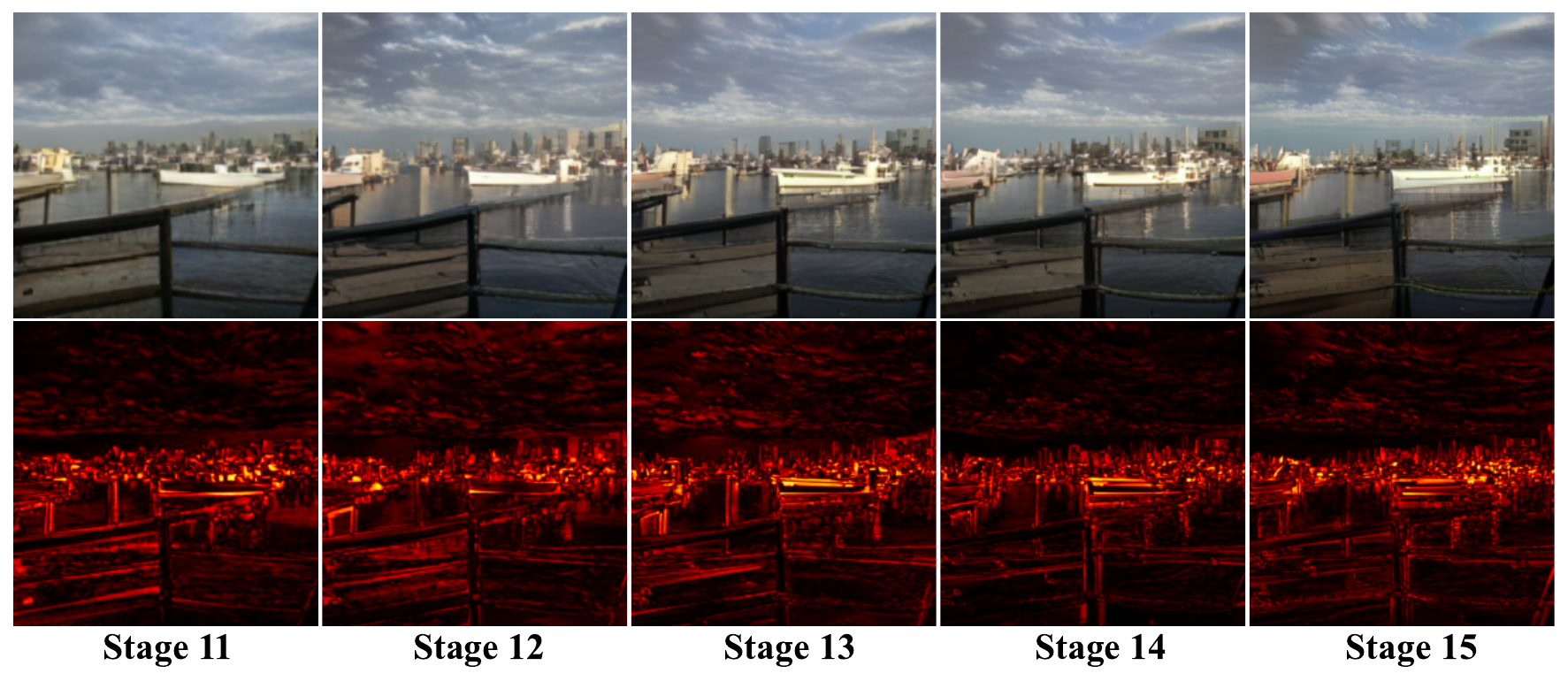}
  \vspace{-0.4cm}
  \caption{$\ell_1$ difference between consecutive stages in FlexVAR. Redder areas indicate greater refinement. Low-frequency regions show minimal change in later stages.}
  \label{fig:flexvar_l1}
\end{figure}

To assess whether the token redundancy observed in high-resolution stages originates from the next-scale prediction scheme itself or from the residual VAE architecture (used in models like HART and Infinity), we evaluate SparseVAR on \textbf{FlexVAR}~\cite{VAR}, a next-scale prediction model that does \emph{not} use residual VAEs. Unlike HART or Infinity, FlexVAR directly predicts the entire token map at each stage without computing residuals. This makes it a suitable testbed for isolating the effect of the next-scale prediction paradigm. As shown in Fig.~\ref{fig:flexvar_l1}, we visualize the $\ell_1$ differences between consecutive stages in FlexVAR. We find that low-frequency regions receive minimal refinement in later stages, while high-frequency regions continue to be updated. This confirms that token redundancy is inherent to the multi-stage refinement design of next-scale prediction, independent of residual VAE components.

We then apply SparseVAR to FlexVAR by skipping token prediction in low-frequency regions and using interpolated values from the previous stage. As shown in Table~\ref{tab:flexvar_sparsevar}, this yields inference acceleration with minimal degradation in image quality. Although the performance gain is smaller—due to FlexVAR’s lower 512$\times$512 resolution and limited scalability—the results confirm that SparseVAR remains effective even in non-residual VAE settings.

\begin{table}[h]
\centering
\caption{Evaluation on ImageNet (512$\times$512) using FlexVAR with and without SparseVAR.}
\vspace{-0.3cm}
\begin{tabular}{lcc}
\toprule
Model & FID $\downarrow$ & Latency (s) $\downarrow$ \\
\midrule
FlexVAR & 4.33 & 0.52 \\
\quad + SparseVAR & 4.38  & 0.43 \\
\bottomrule
\end{tabular}
\label{tab:flexvar_sparsevar}
\end{table}

\section{Comparison with Token Reduction Methods}

\begin{figure}[h]
  \centering
  \includegraphics[width=0.55\linewidth]{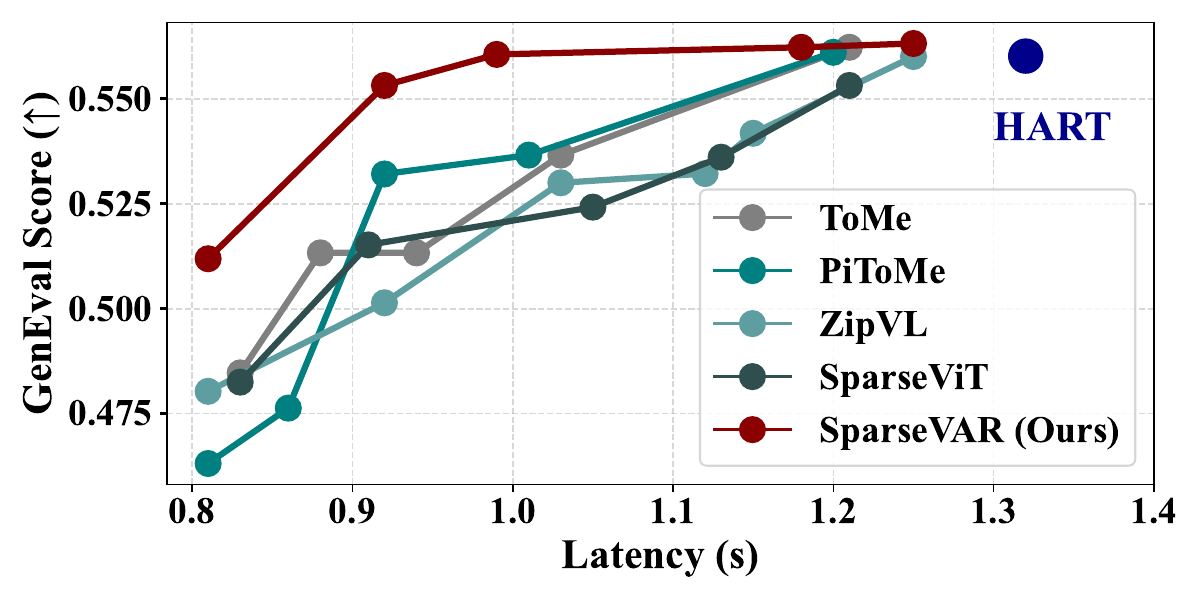}
  \vspace{-0.5cm}
  \caption{Comparison of token reduction methods on GenEval using HART-0.7B.}
  \label{fig:compare_token_methods}
\end{figure}

We compare \methodname{} with representative token reduction strategies, including both token merging and token selection methods, applied during high-resolution stages. For token merging, we evaluate ToMe~\cite{bolya2022tome} and PiToMe (NeurIPS 2024), sweeping sparsity levels from 10\% to 50\%. For token selection, we test SparseViT (CVPR 2023) and ZipVL~\cite{he2024zipvl}, both of which are compatible with FlashAttention. We explore sparsity levels ranging from 10\% to 80\%. As shown in Fig.~\ref{fig:compare_token_methods}, token merging tends to degrade image quality significantly under higher sparsity settings, while selection methods based on attention scores or token norms fail to identify redundancy effectively in next-scale prediction models like VAR. This is largely due to the strong local dependency patterns in these models, which differ from typical vision transformers.

In contrast, \methodname{} introduces a new insight: that redundancy predominantly stems from low-frequency tokens in later high-resolution stages. It leverages a frequency-aware metric (MSE change in high-frequency blocks) to perform dynamic token exclusion, achieving a better balance between quality and computational efficiency.

\section{Performance on Complex Scenes}

\begin{figure}[h]
  \centering
  \includegraphics[width=0.33\linewidth]{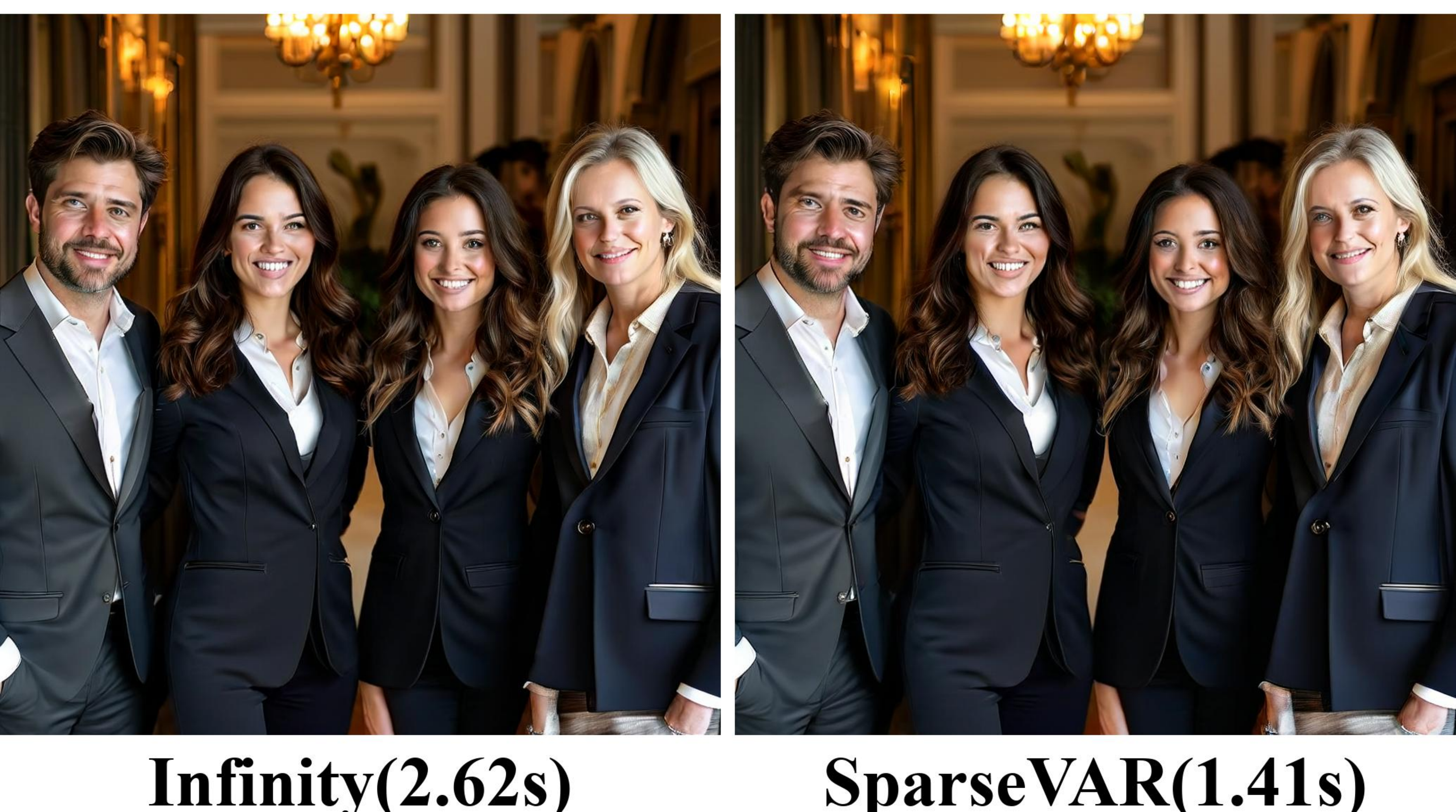}
  \includegraphics[width=0.33\linewidth]{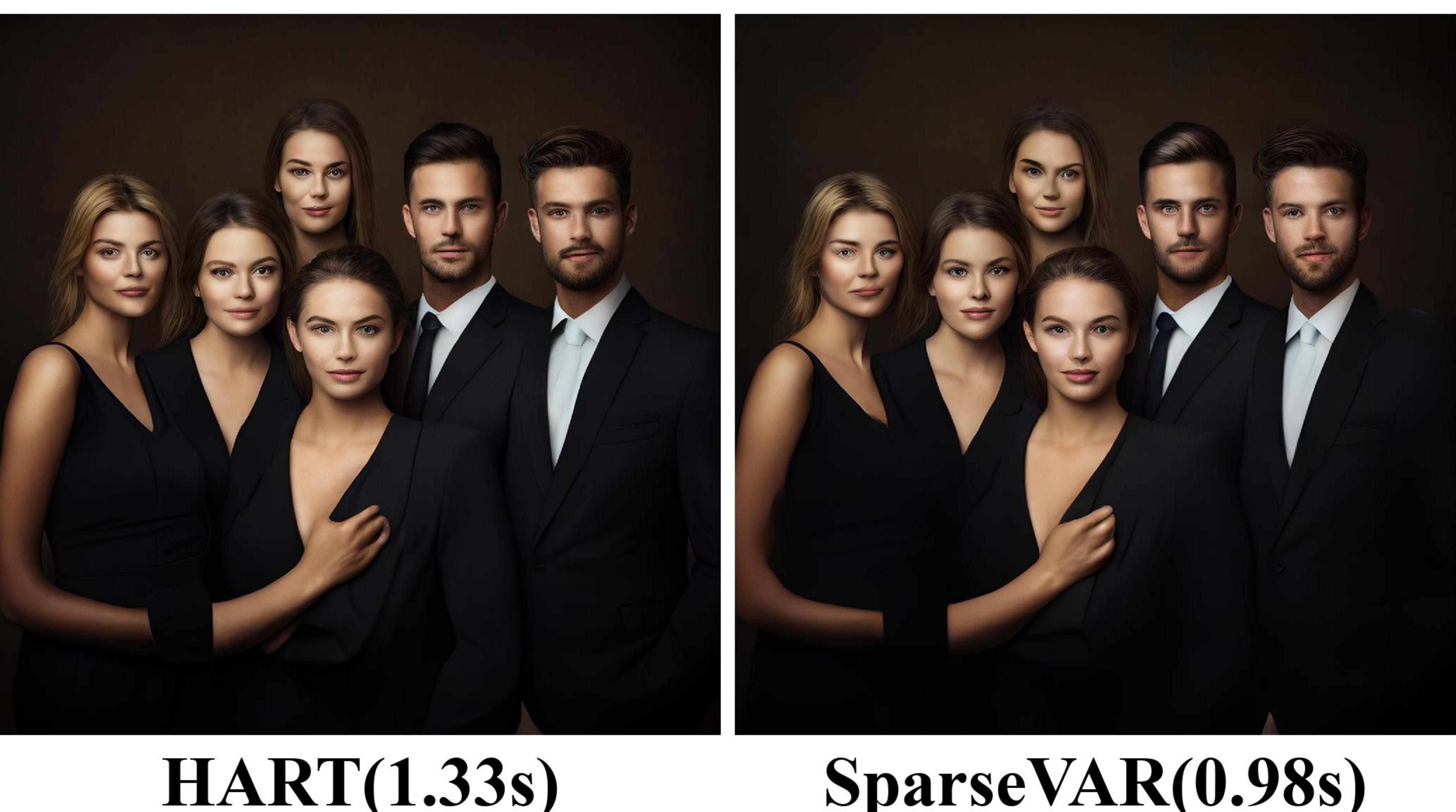}
  \includegraphics[width=0.33\linewidth]{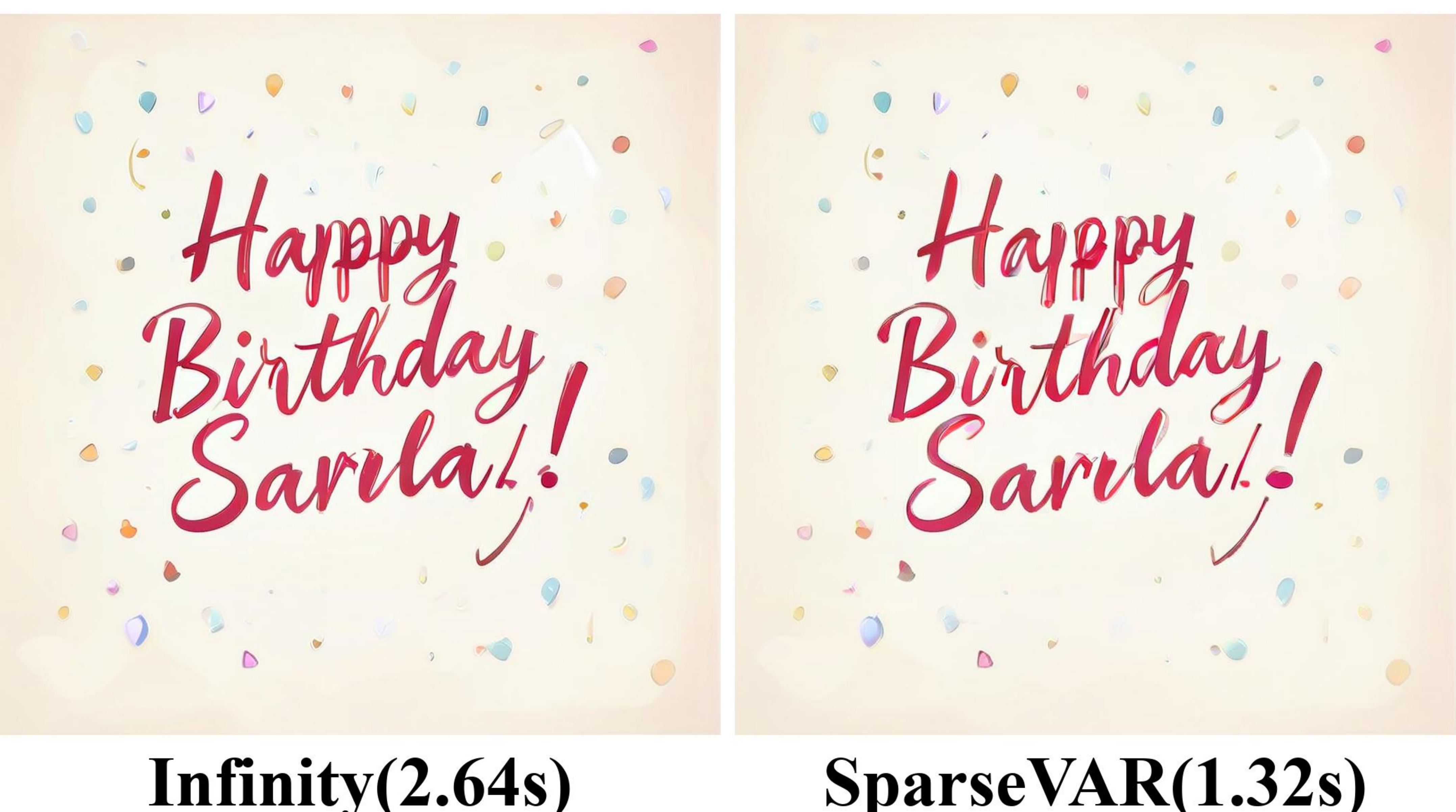} \\
  \includegraphics[width=0.33\linewidth]{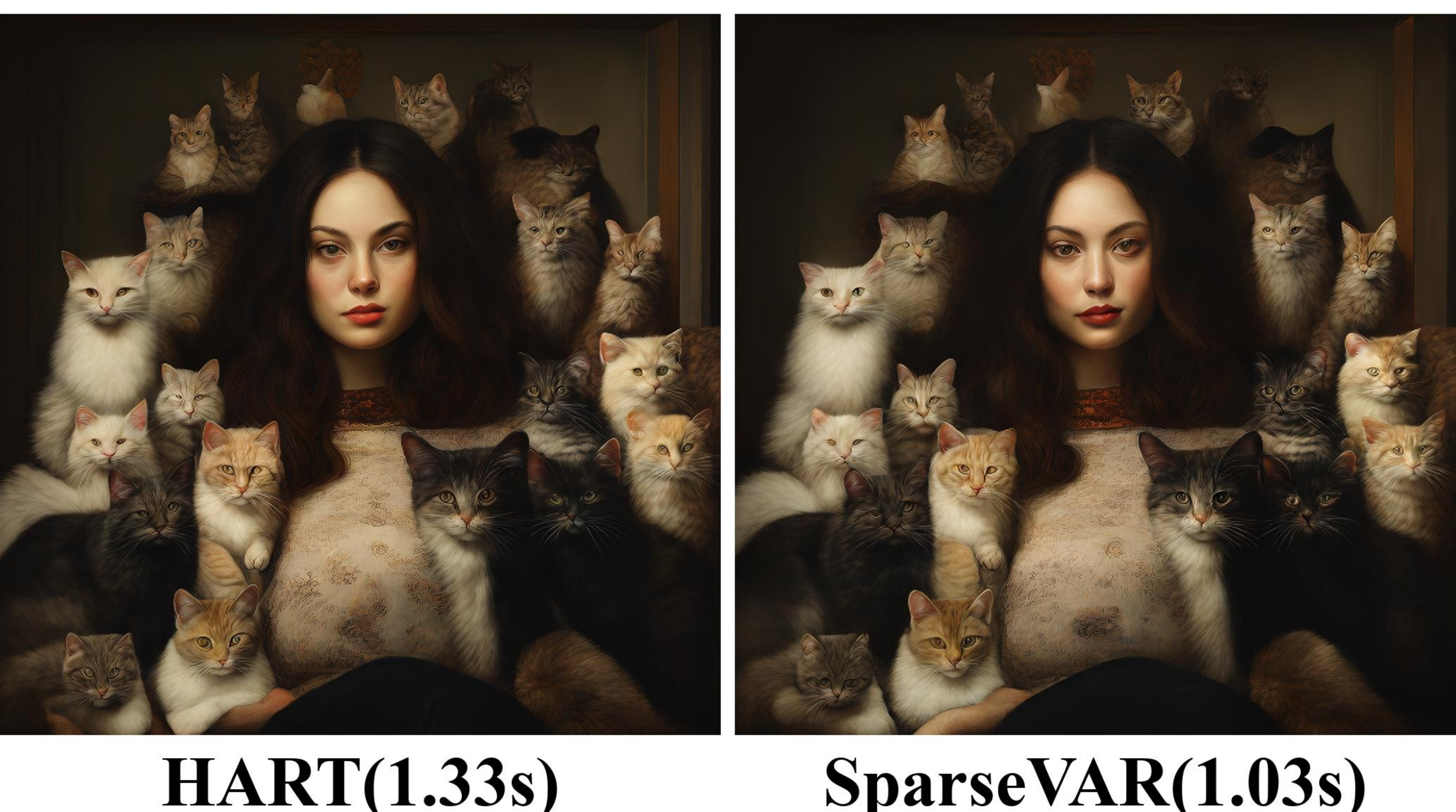}
  \includegraphics[width=0.33\linewidth]{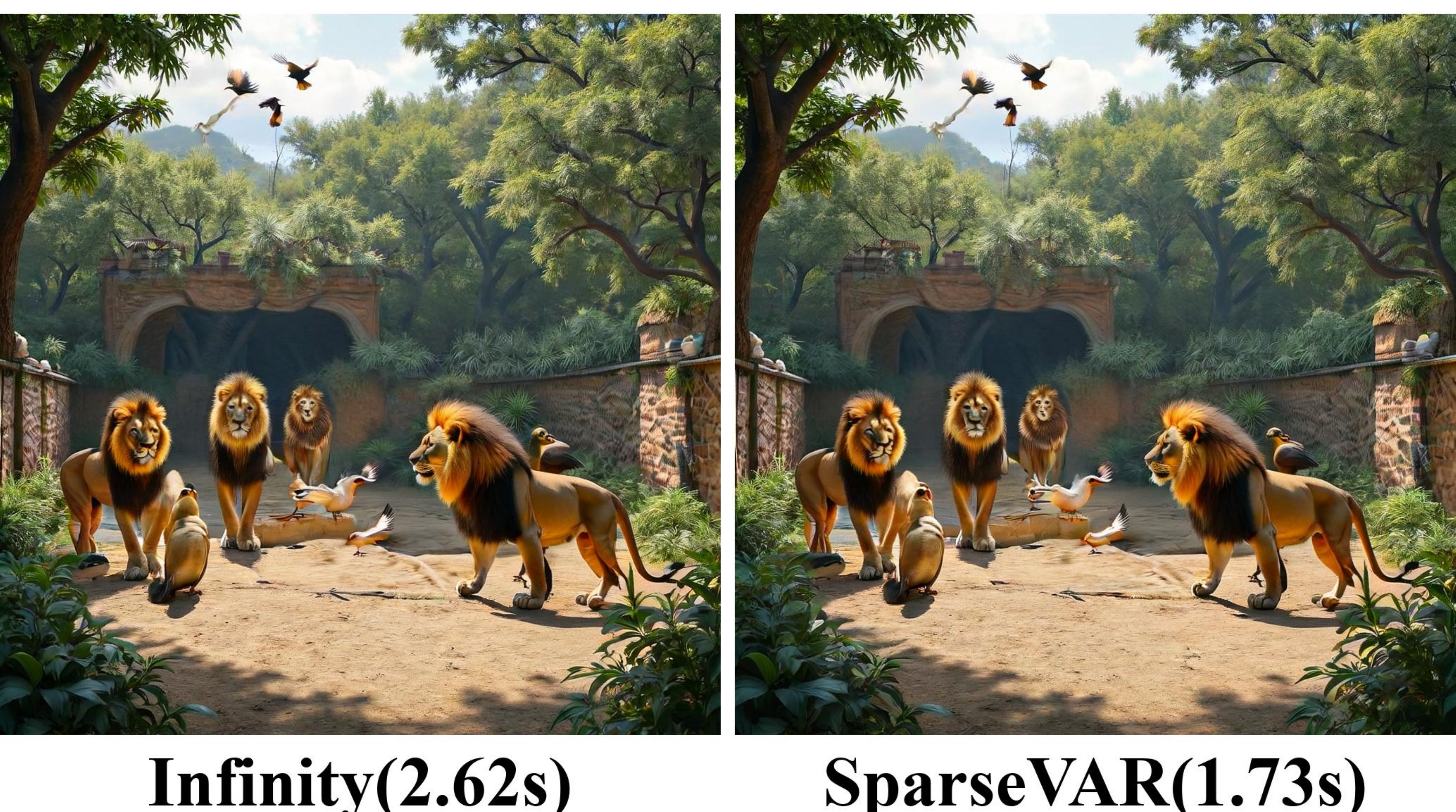}
  \includegraphics[width=0.33\linewidth]{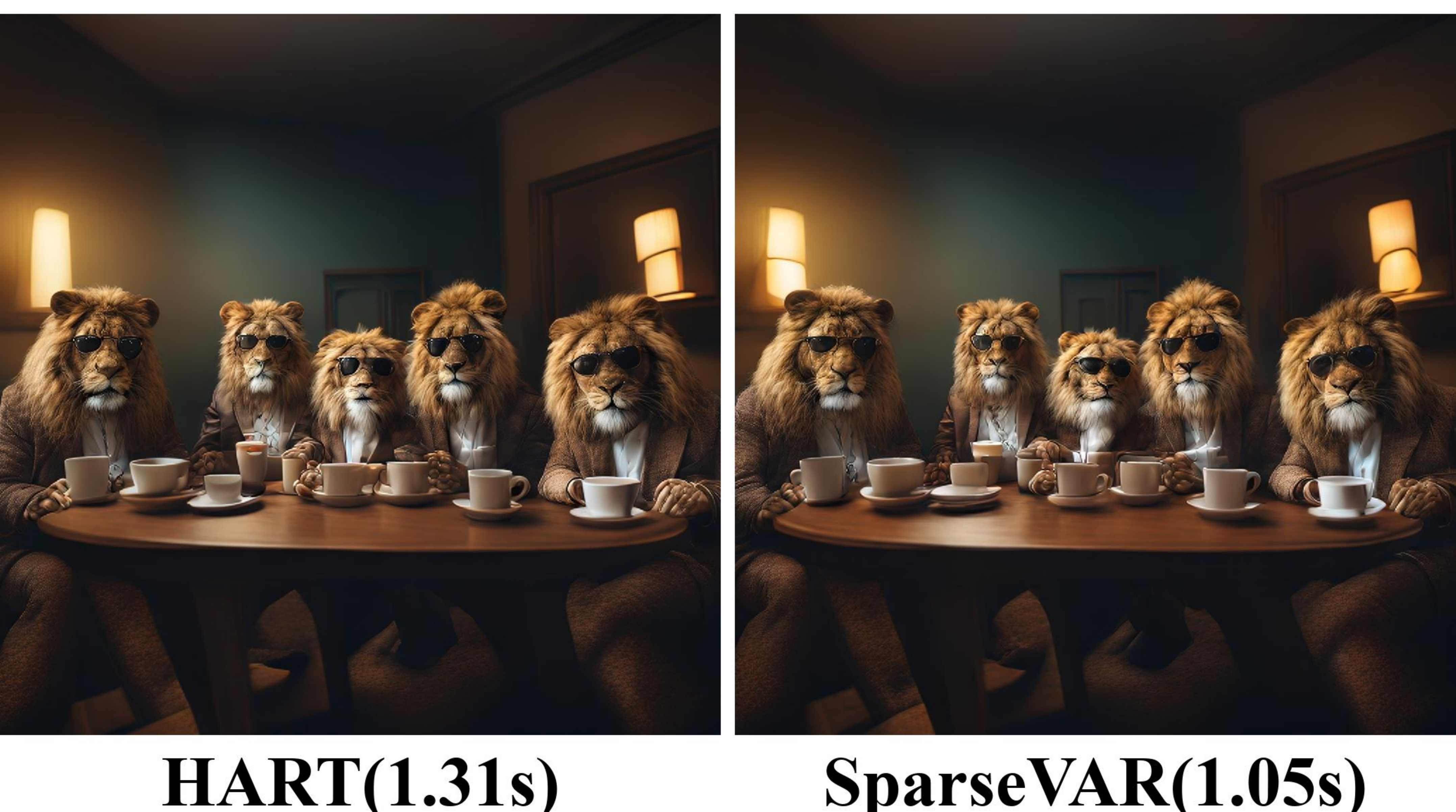} \\
  
  \vspace{-0.3cm}
  \caption{Visualization results on complex scenes with multiple subjects and high-frequency details (e.g., text, faces, groups). SparseVAR maintains visual quality while providing acceleration.}
  \label{fig:complex_visuals}
\end{figure}

To further evaluate the robustness of \methodname{} in real-world settings, we conduct qualitative analysis on complex scenes involving multiple subjects and high-frequency content. These include images with crowded human figures, multiple animal faces, textual elements, and fine-grained visual structures. As shown in Fig.~\ref{fig:complex_visuals}, \methodname{} is able to maintain high-fidelity generation quality even in these more challenging scenarios. Despite early exiting in low-frequency regions, critical high-frequency areas such as facial features and text remain well-preserved. These results suggest that \methodname{} generalizes beyond simple single-object scenes, and remains effective under diverse and complex image distributions.

\section{More Experimental Details}
\subsection{Implementation Details of Figure~6}

Assume that tokens in the $k$-th stage ($k \geq P$) are candidates for early exclusion. For logits, the logits from the $(k-1)$-th stage are interpolated to match the resolution of the $k$-th stage. The average cosine similarity between each token and its neighboring $3 \times 3$ region is computed. Tokens are classified as low-frequency and marked for early exclusion if their neighboring similarity exceeds $\tau$ times the maximum observed similarity across all tokens, where $\tau \in [0.2, 0.3, 0.4, 0.5, 0.6, 0.7]$. For the $\ell_1$ metric, the cumulative token maps from the $(k-1)$-th and $(k-2)$-th stages are passed through the VAE decoder to generate images. The $\ell_1$ difference between these two images is then computed. Tokens are classified as low-frequency and marked for early exclusion if their $\ell_1$ difference is below $\tau$ times the maximum observed $\ell_1$ difference across all tokens, where $\tau \in [0.04, 0.05, 0.06, 0.07, 0.1, 0.2, 0.3]$. For the MSE metric, $\tau \in [0.2, 0.3, 0.4, 0.5, 0.6, 0.7]$.

\section{More Visualizations of Observations}

This subsection provides additional visualization examples corresponding to the observations discussed in the main text.

\subsection{Minimal Impact of High-Resolution Residuals on Low-Frequency Regions}
\begin{figure}[h]
\centering
\begin{subfigure}[b]{0.95\textwidth}
    \centering
    \includegraphics[width=0.95\linewidth]{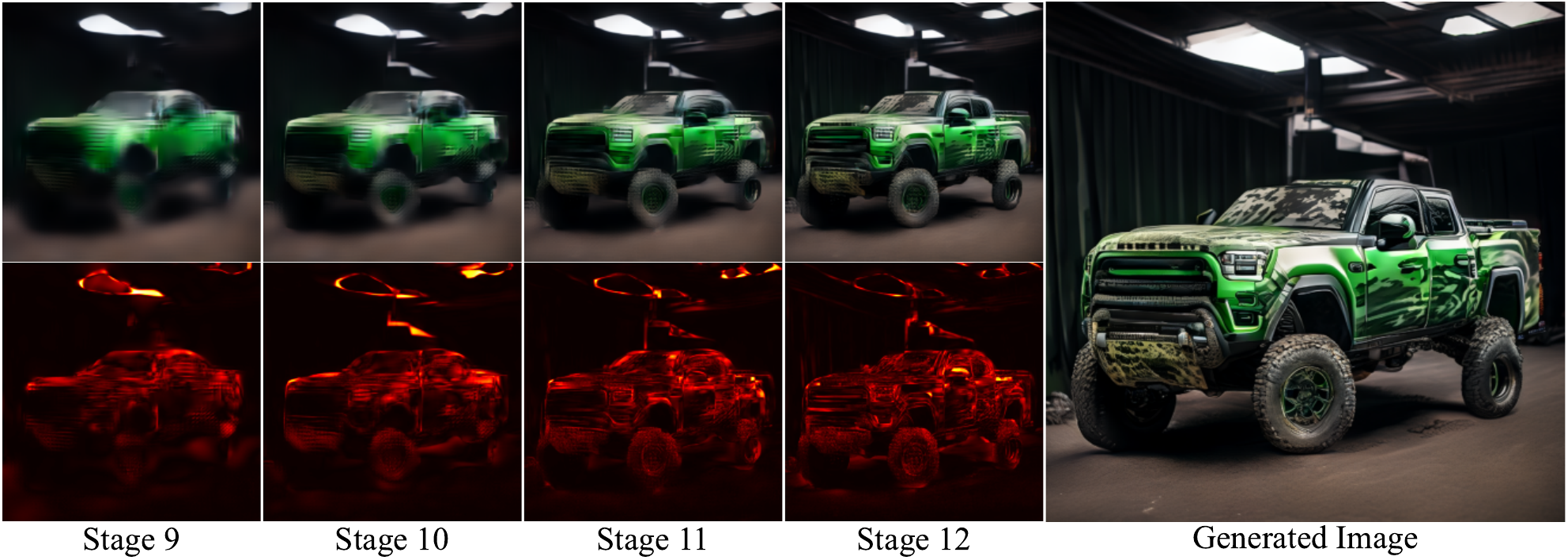}
\end{subfigure}
\vspace{-0.4cm}
\caption{More visualizations of HART-0.7B.}
\label{fig:sup_motivation1_hart}
\end{figure}

\begin{figure}[h]
\centering
\begin{subfigure}[b]{0.95\textwidth}
    \centering
    \includegraphics[width=0.95\linewidth]{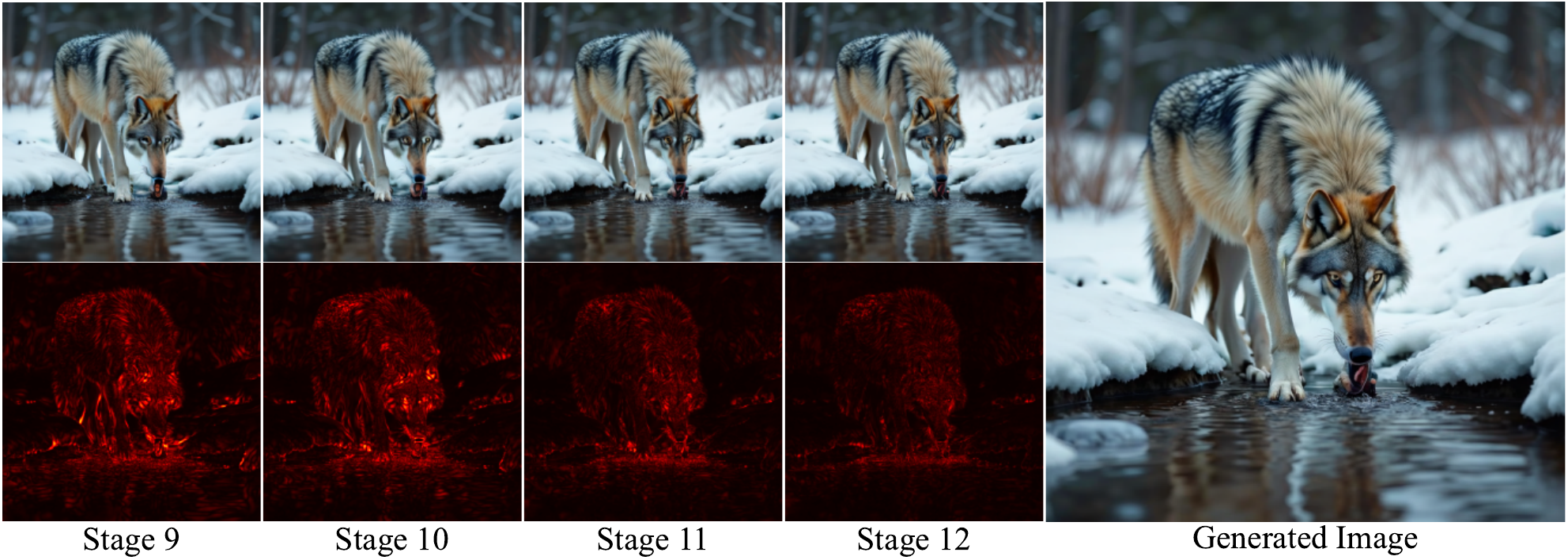}
\end{subfigure}
\vspace{-0.4cm}
\caption{More visualizations of Infinity-2B.}
\label{fig:sup_motivation1_infinity}
\end{figure}

In this section, we present additional visualizations of the $\ell_1$ changes in the generated images of the accumulated token maps between each of the last four stages and the previous stage. As shown in Figures~\ref{fig:sup_motivation1_hart} and~\ref{fig:sup_motivation1_infinity}, it is evident that the feature map $r_k$ generated in high-resolution stages have a limited impact on the low-frequency regions in both HART and Infinity.

\subsection{Distinct Regional Focus of Blocks in Next-Scale Prediction Models}

\begin{figure}[h]
\centering
\begin{subfigure}[b]{0.99\textwidth}
    \centering
    \includegraphics[width=0.46\linewidth]{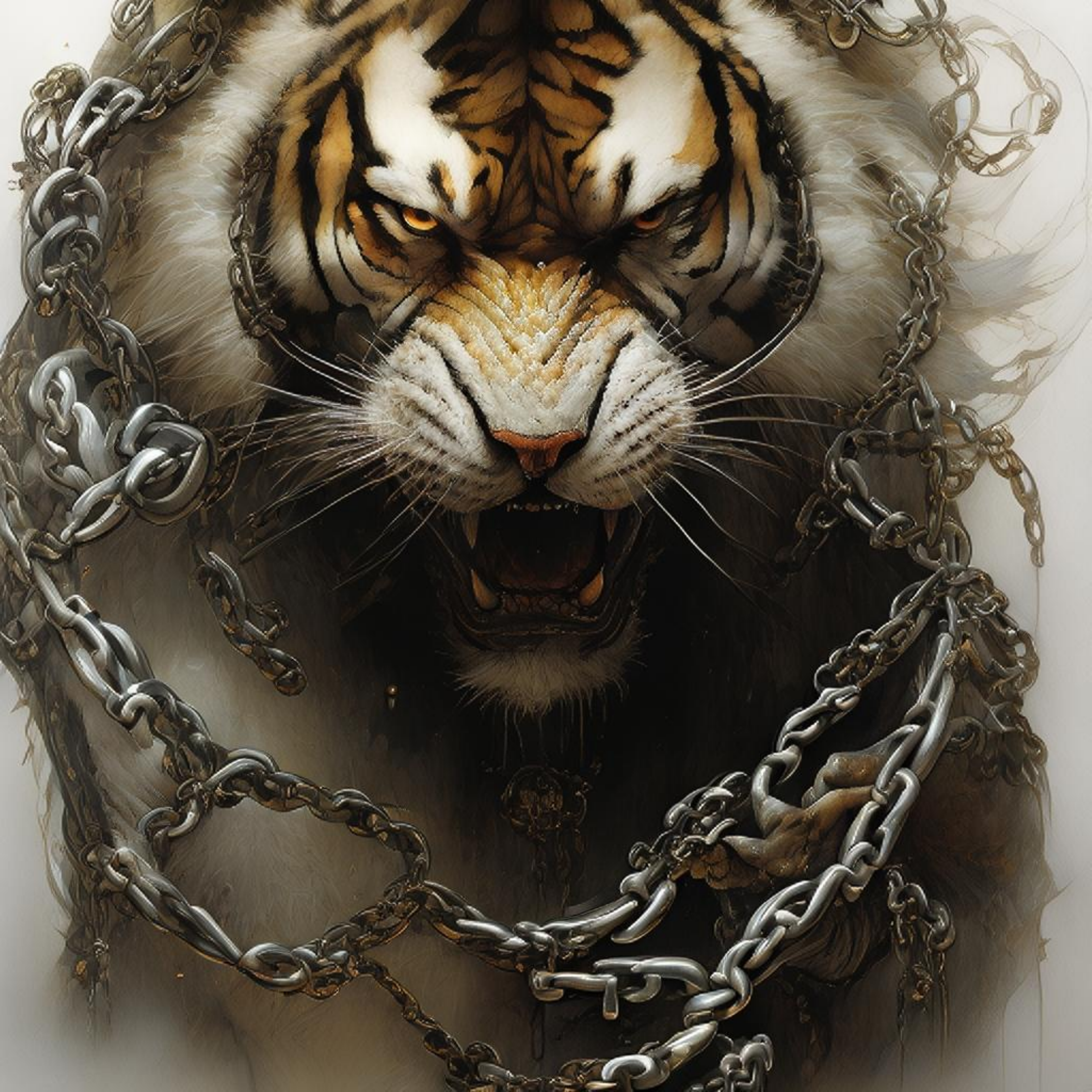}
    \includegraphics[width=0.46\linewidth]{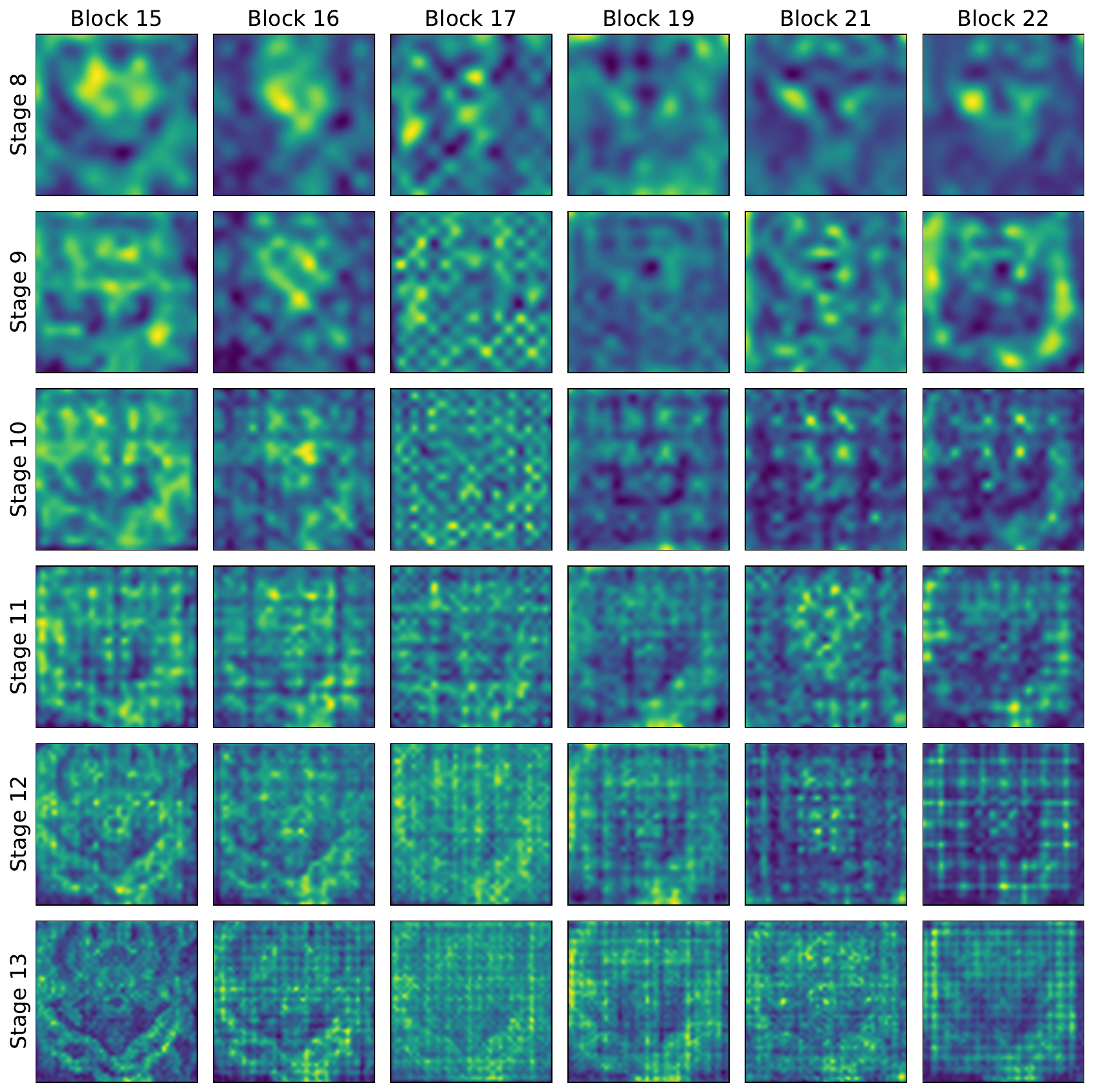}
\end{subfigure}
\vspace{-0.4cm}
\caption{More visualizations of HART-0.7B.}
\label{fig:sup_motivation2_1}
\end{figure}

\begin{figure}[h]
\centering
\begin{subfigure}[b]{0.99\textwidth}
    \centering
    \includegraphics[width=0.46\linewidth]{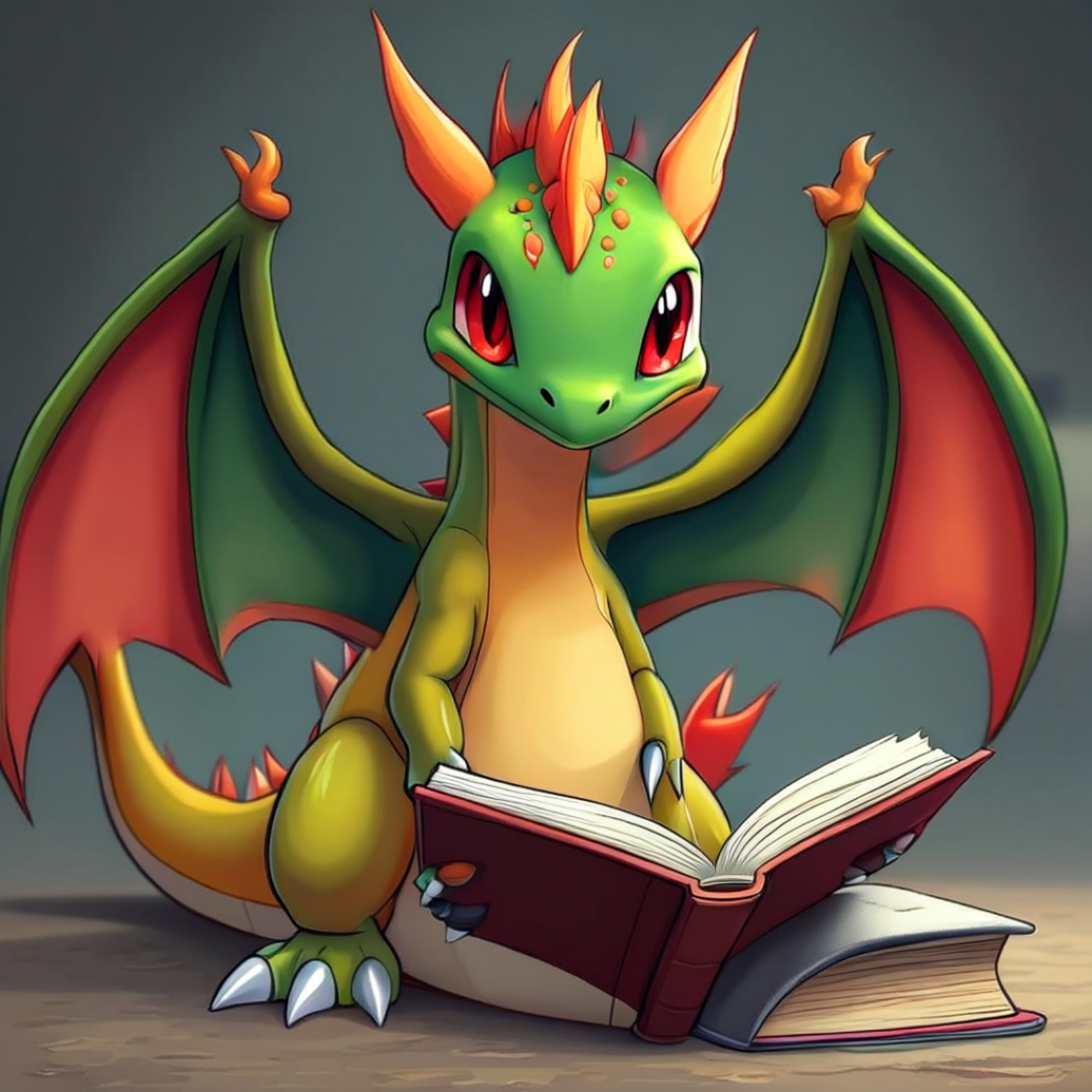}
    \includegraphics[width=0.46\linewidth]{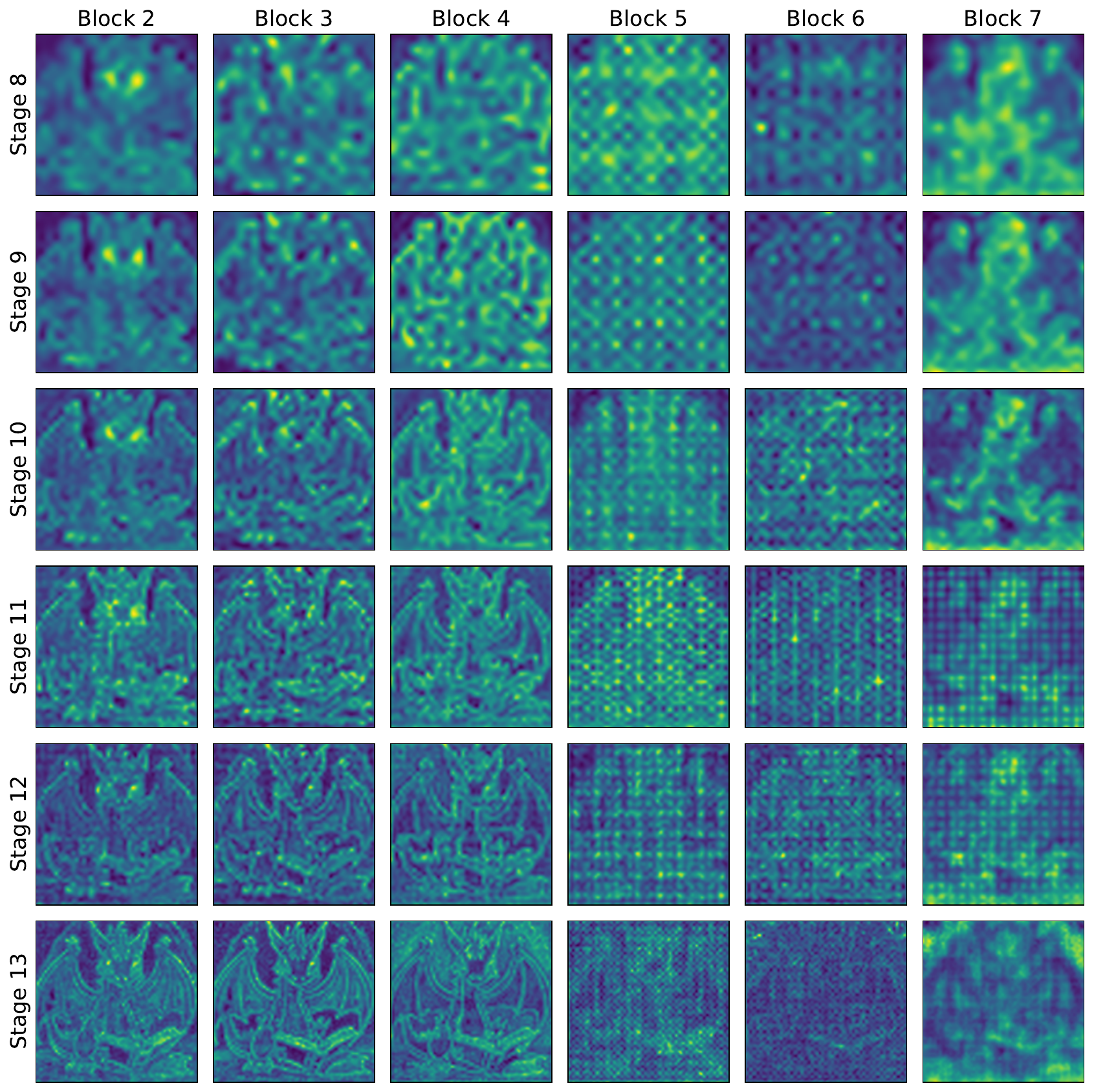}
\end{subfigure}
\vspace{-0.4cm}
\caption{More visualizations of Infinity-2B.}
\label{fig:sup_motivation2_2}
\end{figure}

In this section, we present more detailed visualizations of the MSE changes for some blocks in the final five stages. As shown in Figure~\ref{fig:sup_motivation2_1} and Figure~\ref{fig:sup_motivation2_2}, it is evident that different blocks focus on distinct regions, with some emphasizing high-frequency regions and others targeting low-frequency regions. Based on the most effective clarification, we select the 3rd block for Infinity-2B and the 16th block for HART-0.7B.

\subsection{High Logits Similarity in Certain Low-Frequency Regions}

Low-frequency regions refer to areas in the image with slow variations, where an intuitive hypothesis is that the predicted residuals for these regions at high-resolution stages should be relatively similar. To explore the similarity of predicted residuals in low-frequency regions, we visualize the average cosine similarity of logits between each token and its neighboring 3$\times$3 region at stages 8-11 of HART and Infinity. As shown in Figure~\ref{fig:sup_motivation3_1} and Figure~\ref{fig:sup_motivation3_2}, for certain low-frequency regions, the logits predicted at each token position exhibit high similarity with those of neighboring tokens. Based on this observation, \textit{we propose retaining a subset of anchor tokens to represent the predictions of their neighboring regions, thereby improving the generation quality of low-frequency areas.}



\begin{figure}[h]
\centering
\begin{subfigure}[b]{0.91\textwidth}
    \centering
    \includegraphics[width=0.99\linewidth]{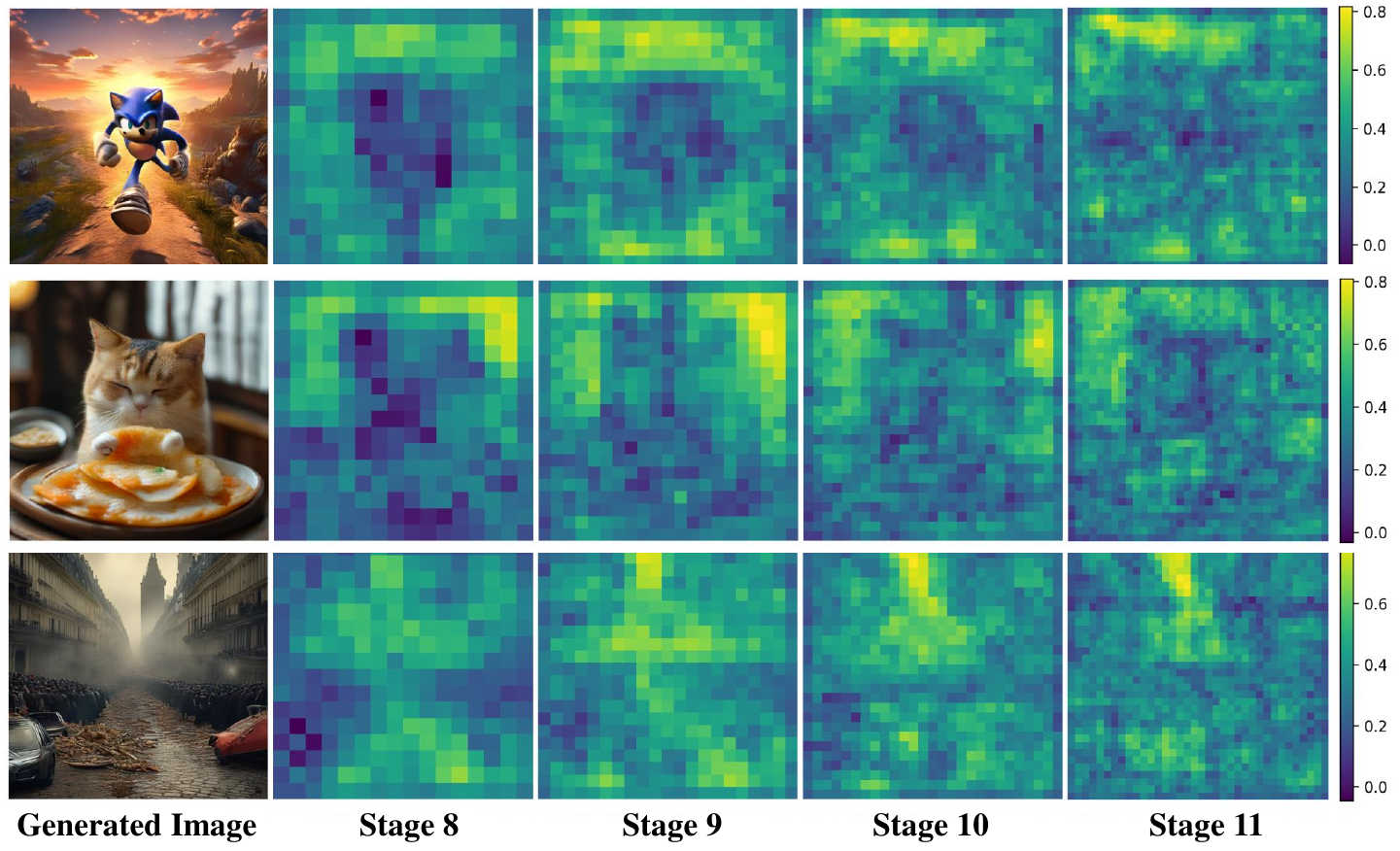}
    \label{fig:sup_motivation3_1}
\end{subfigure}
\vspace{-0.4cm}
\caption{More visualizations of HART-0.7B.}
\label{fig:sup_motivation3_1}
\end{figure}

\vspace{-0.4cm}

\begin{figure}[h]
\centering
\begin{subfigure}[b]{0.91\textwidth}
    \centering
    \includegraphics[width=0.99\linewidth]{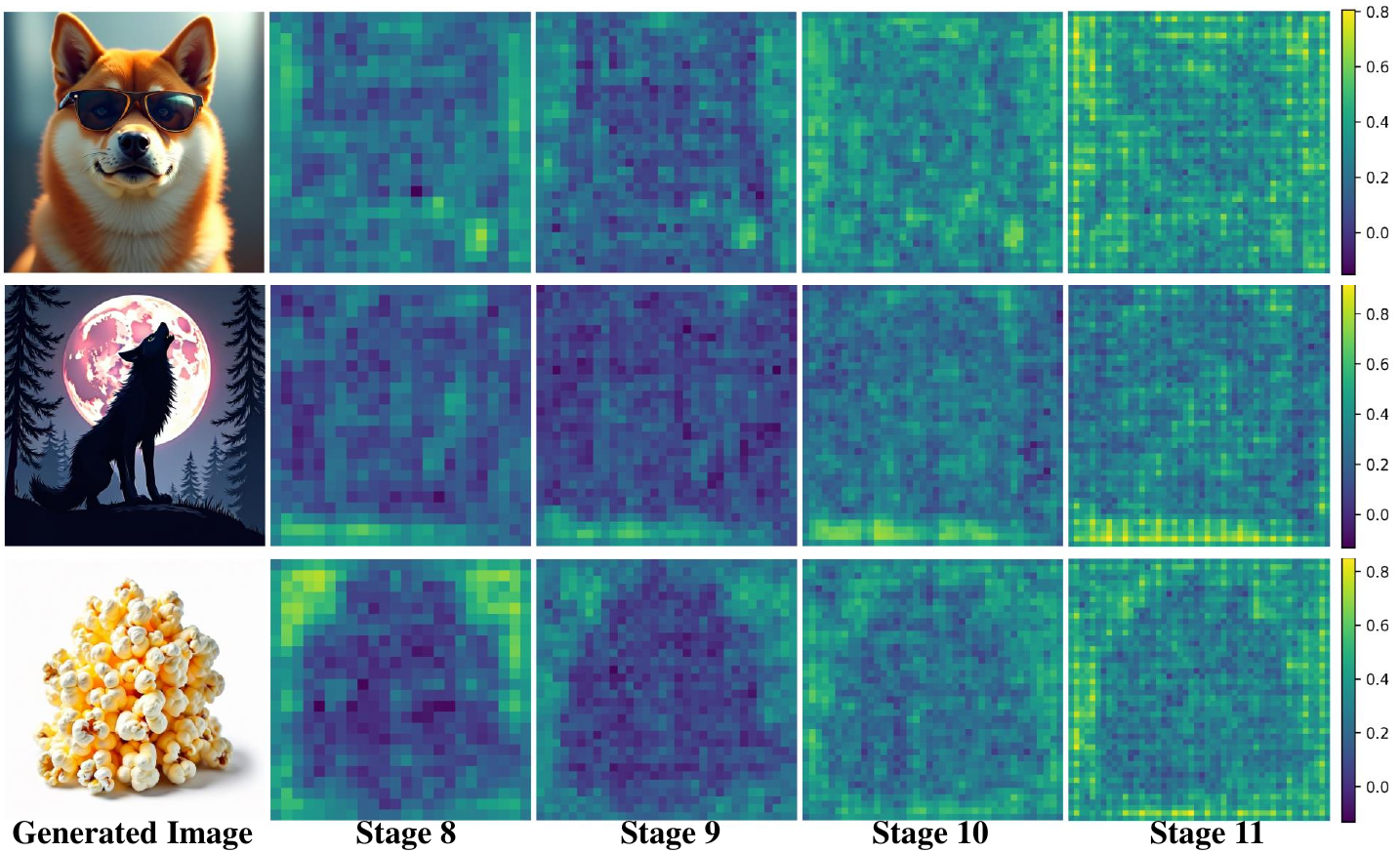}
    \label{fig:sup_motivation3_2}
\end{subfigure}
\vspace{-0.4cm}
\caption{More visualizations of Infinity-2B.}
\label{fig:sup_motivation3_2}
\end{figure}

\section{Visualizations of Attention Maps in Next-Scale Prediction}

We visualize the attention maps of three blocks from HART-0.7B across the last four stages. As shown in the figure, the next-scale prediction model exhibits high attention scores for nearly all tokens with respect to their neighboring tokens, while the attention scores for more distant tokens are close to zero. 

\begin{figure*}[h]
\centering
\begin{subfigure}[b]{0.99\textwidth}
    \centering
    \includegraphics[width=0.95\linewidth]{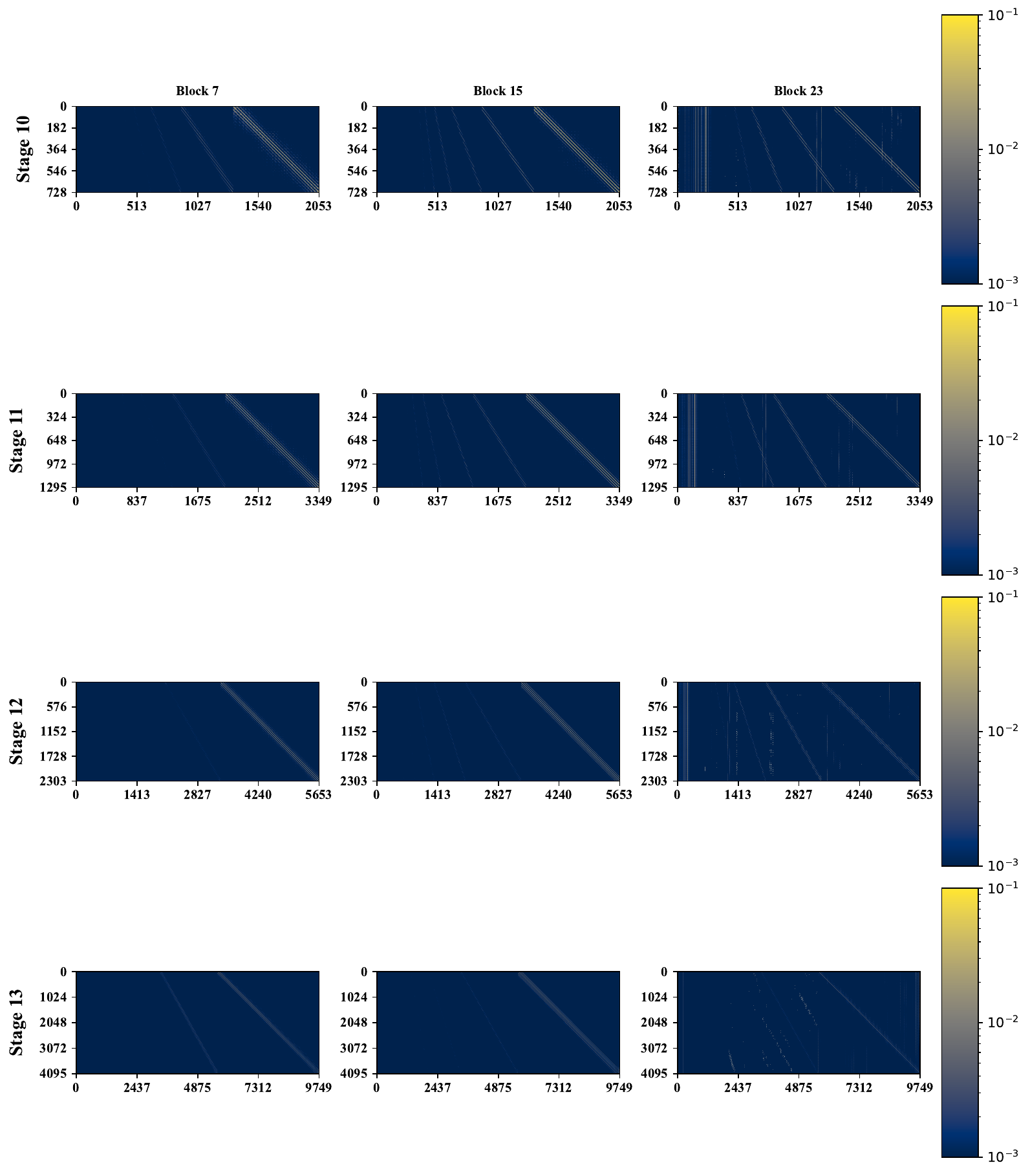}
    \label{fig:attn_maps}
\end{subfigure}
\vspace{-0.4cm}
\caption{Visualization of attention maps of \methodname. The x-axis represents the indices of all tokens in the current stage and the KV cache, while the y-axis represents the token indices in the current stage.}
\label{fig:visualizations_sup}
\end{figure*}

\clearpage
\section{More Qualitative Visualizations}

\begin{figure*}[h]
\centering
\begin{subfigure}[b]{0.99\textwidth}
    \centering
    \includegraphics[width=0.99\linewidth]{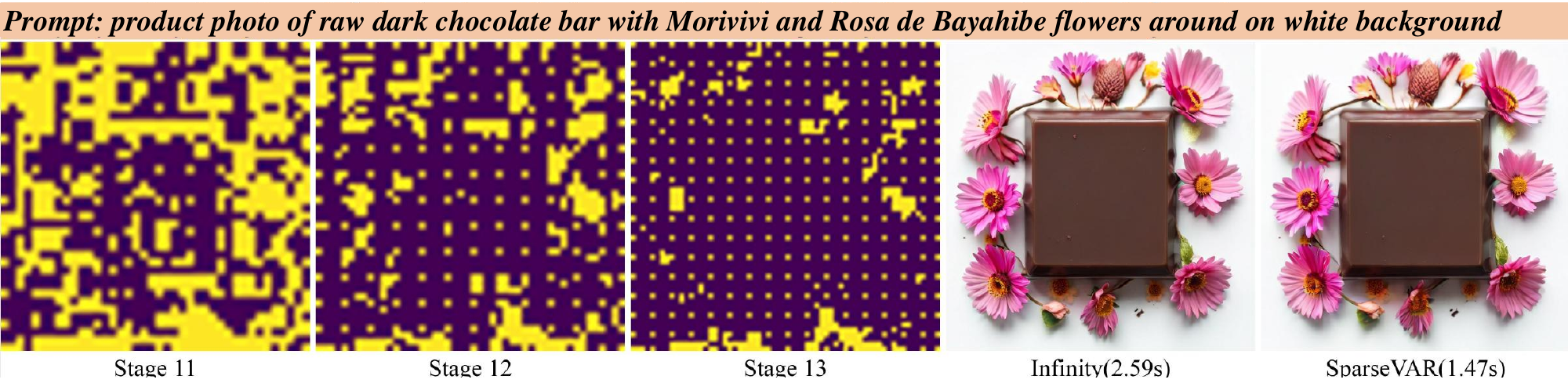}
    \label{fig:vis_sparsity_infinity_2}
\end{subfigure}
\begin{subfigure}[b]{0.99\textwidth}
    \centering
    \includegraphics[width=0.99\linewidth]{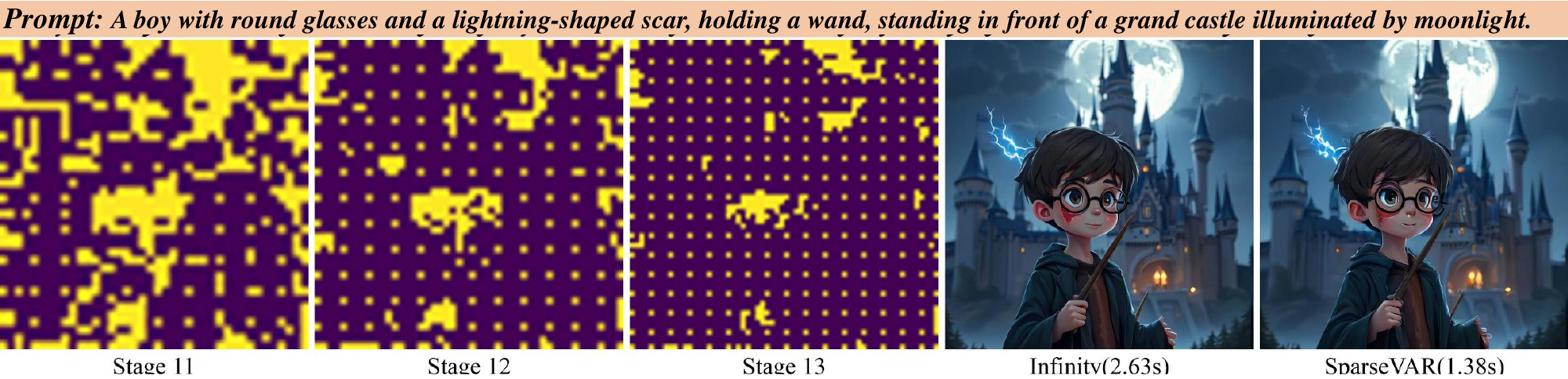}
    \label{fig:vis_sparsity_infinity_4}
\end{subfigure}
\begin{subfigure}[b]{0.99\textwidth}
    \centering
    \includegraphics[width=0.99\linewidth]{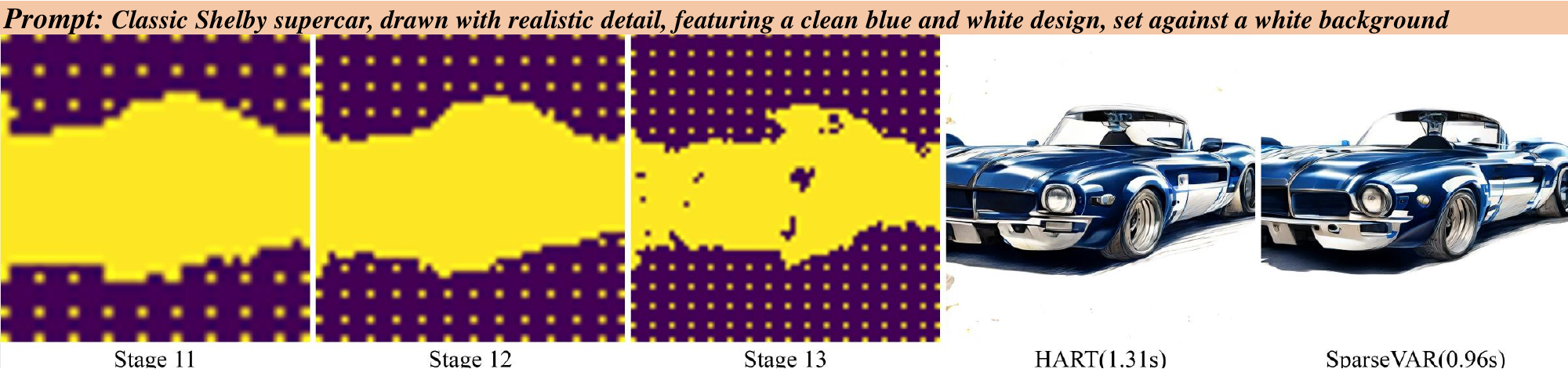}
    \label{fig:vis_sparsity_hart_3}
\end{subfigure}
\begin{subfigure}[b]{0.99\textwidth}
    \centering
    \includegraphics[width=0.99\linewidth]{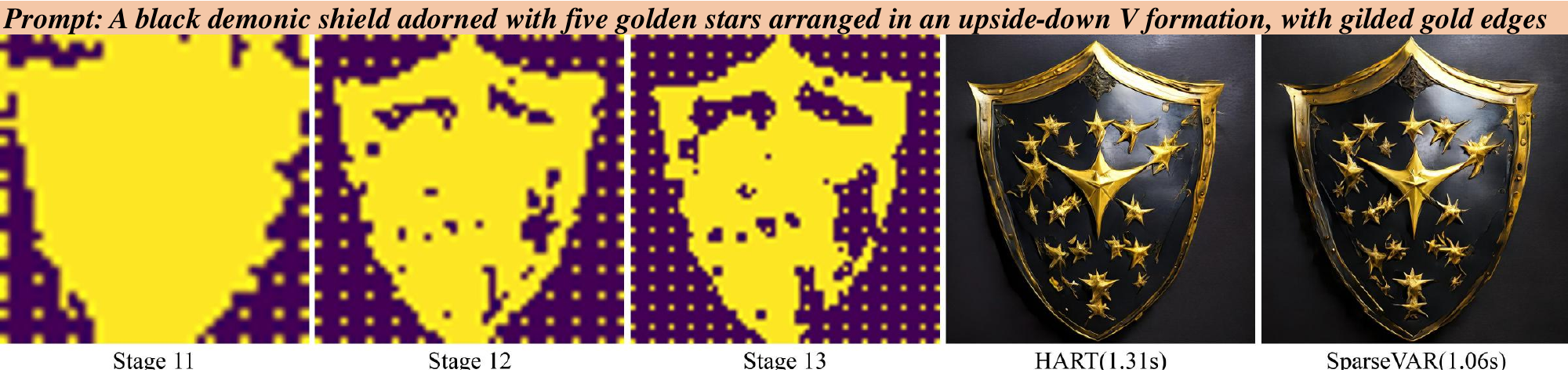}
    \label{fig:vis_sparsity_hart_4}
\end{subfigure}
\vspace{-0.4cm}
\caption{Qualitative visualizations of \methodname.}
\label{fig:visualizations_sup}
\end{figure*}

\end{document}